\documentclass{article} 
\usepackage{iclr2027_conference,times}

\usepackage{amsmath,amsfonts,bm}

\def\eqref#1{equation~\ref{#1}}

\def\1{\bm{1}}

\DeclareMathAlphabet{\mathsfit}{\encodingdefault}{\sfdefault}{m}{sl}
\SetMathAlphabet{\mathsfit}{bold}{\encodingdefault}{\sfdefault}{bx}{n}

\usepackage{hyperref}
\usepackage{cleveref}
\usepackage{url}
\usepackage[nointegrals]{wasysym}
\usepackage{booktabs}
\usepackage{graphicx}
\usepackage{amssymb}
\usepackage{enumitem}
\usepackage[table]{xcolor}
\usepackage{fvextra}
\usepackage{subcaption}

\definecolor{pairregion}{HTML}{EEF3F9}

\newcommand{\pairshade}{%
  \raisebox{0.05ex}{%
    \fcolorbox{black!30}{pairregion}{\rule{0pt}{0.5em}\rule{0.8em}{0pt}}%
  }%
}
\newcommand{\artifacturl}[0]{\url{https://anonymous.4open.science/r/PAIR-release-3BEB}}

\title{PAIR: Memory-Efficient and Fast Inference with Reading–Reasoning Fusion}

\author{%
Sejun Park \\
Siebel School of Computing and Data Science \\
University of Illinois Urbana-Champaign \\
Urbana, IL 61801, USA \\
\texttt{sejun3@illinois.edu}
\And
Hancheng Xia \\
Siebel School of Computing and Data Science \\
University of Illinois Urbana-Champaign \\
Urbana, IL 61801, USA \\
\texttt{hxia7@illinois.edu}
\And
Yongjoo Park \\
Siebel School of Computing and Data Science \\
University of Illinois Urbana-Champaign \\
Urbana, IL 61801, USA \\
\texttt{yongjoo@illinois.edu}
}

\iclrfinalcopy 
\begin{document}

\maketitle
\fancyhead{}
\renewcommand{\headrulewidth}{0pt}

\begin{abstract}
For context-based question answering, using one large model for both evidence extraction and answer reasoning requires substantial memory and computation.
Existing approaches trade accuracy for efficiency within a single model only.
We propose a multi-model approach for efficient inference: our method, PAIR, employs two differently sized models for evidence extraction and answer reasoning, respectively.
A smaller Reader model processes and retains the full context, while a larger Reasoner model sees only the question and generated sequence without directly attending to the context.
These capabilities are fused at the logit level: at each token, both Reader and Reasoner produce a prediction, and their logits are combined into one, behaving as a single next-token-prediction model. Fine-tuning further boosts their performance.
On HotpotQA, PAIR with a 32B Reasoner retains 96\% of that same 32B model’s full-context accuracy while using roughly one quarter of its KV memory. At 45.7K-token contexts, PAIR achieves up to 3.2$\times$ higher end-to-end throughput than the full-context 32B model on the same GPU. PAIR also improves the accuracy-memory and accuracy-throughput trade-offs over evaluated token-reduction baselines. Its accuracy-preserving behavior generalizes across Qwen2.5, OLMo-3, and Gemma-3 model families. 

\end{abstract}

\section{Introduction}
To produce correct answers from a given context, a language model
requires two critical capabilities: \emph{extracting} evidence and \emph{reasoning} toward an answer~\citep{rag,hotpotqa}.
\emph{Evidence extraction} identifies information in the context that is relevant to the question; \emph{answer reasoning} combines that evidence to generate the answer.
While a larger model tends to produce more accurate answers, a larger model also incurs
(i) higher memory consumption for retaining the KV cache of the context, e.g., documents~\citep{lazyattention} or past conversation history~\citep{slotguard}, and 
(ii) longer latency from processing a larger number of model parameters. 
A technique that can significantly reduce memory consumption and latency, at negligible accuracy loss, would be desirable in resource-constrained environments.


Existing work has explored accuracy-efficiency trade-off \textbf{within a single model}. 
First, token selection and KV eviction reduce the amount of context processed or retained in the KV cache~\citep{specprefill,snapkv,cokv,expectedattention}.
While these methods offer straightforward efficiency gains, their accuracy drops sharply when informative tokens are dropped.
Second, KV quantization reduces memory consumption by converting the original high-precision KV representations into lower-precision ones, at the cost of accuracy~\citep{KIVI,kvquant}. 
However, this yields little latency improvement, since the original high-precision representation must still be computed before quantization.

Our core idea is to achieve accuracy-efficiency trade-off by \textbf{separating evidence extraction and answer reasoning into differently sized models}.
A small model processes and retains the full context for evidence extraction.
A large model focuses on answer reasoning: 
once the relevant evidence appears in the generated sequence, the larger model uses it without directly attending to the full context.
Section~\ref{sec:branch_motivation} gives a concrete example.
Since only the small model processes the context, both memory consumption and latency are significantly lower than when using the large model alone.
The large model remains fully involved during answer reasoning, thus retaining near-original accuracy.
The additional memory and computational overhead from running the small model is minor compared to the efficiency gains described above.

We call our approach \emph{\textbf{P}arallel \textbf{A}symmetric \textbf{I}nference with \textbf{R}eader-Reasoner Separation} (PAIR). The Reader (a small model) sees the context and question, while the Reasoner (a large model) sees only the question and generated sequence. At each decoding step, the Reader and Reasoner each produce a next-token prediction (in logits). 
Those logits are fused using a weighted sum to produce the next token, which is appended to the generated sequence and becomes visible to both models from the next decoding step.
Since the Reader continues to see the original context throughout decoding, it can write extracted evidence in the generated sequence, making it naturally visible to the Reasoner.
Even without explicit coordination, this logit fusion between the two allows seamless evidence extraction and answer reasoning, producing a coherent response with an accurate answer.

To make Reader-Reasoner fusion effective, we investigate two questions.
\textbf{(Generalizable fusion)} Is there an effective logit fusion method that generalizes across a wide range of benchmarks without overfitting?
    We provide a theoretical analysis explaining why a strikingly simple approach---a weighted linear sum with a fixed, training-free constant---can be highly effective.
\, \textbf{(Role specialization)} Can the Reader and Reasoner become more effective at their own roles through additional training?
    We fine-tune both the Reader and Reasoner through a specialized training pipeline.
    This training uses no benchmark-specific labeled data, allowing the same trained models to generalize across various benchmarks we consider.


Our contributions are:
\begin{itemize}[leftmargin=4mm,topsep=-1mm,itemsep=0mm]
    \item We propose that a language model's core function---next token prediction---can be performed by assigning different roles (e.g., evidence extraction and answer reasoning) to different sub-models, while their predictions are fused at the logit level without rearchitecting their internals.

    \item We realize the above role-based collaboration through PAIR, a training-free weighted logit fusion of the Reader and Reasoner and sequential role-specific adaptation. The Reader is first tuned for evidence extraction, after which the Reasoner is tuned to improve the joint prediction.

    \item We demonstrate that PAIR substantially improves the accuracy-efficiency trade-off over the compared baselines. It preserves most of the full-context large model's accuracy while using substantially less KV memory and achieving higher throughput.

\end{itemize}

\section{Related Work}


\paragraph{Memory-efficient inference.}

Existing approaches on memory-efficient inference can be grouped into two broad families. The first reduces the amount of context processed or retained by the model, which can remove information needed for the current answer or later questions over the same context.
Some methods reduce the retained context after prefill through KV eviction.
H$_2$O and SnapKV select states using observed attention patterns, PyramidKV allocates different KV budgets across layers, and Expected Attention estimates importance from future-query distributions~\citep{h2o,snapkv,pyramidkv,expectedattention}.
KVzip performs query-agnostic eviction to support reuse of the compressed KV cache across multiple queries~\citep{kvzip}.
Other methods reduce the context before or during prefill.
LLMLingua and LongLLMLingua compress the input itself, while SpecPrefill uses a lightweight model to select tokens for the larger model's prefill~\citep{llmlingua,longllmlingua,specprefill}.

The second family retains all context positions but reduces the precision of their KV representations, lowering KV memory while leaving full-context prefill computation unchanged. KIVI and KVQuant quantize retained KV states while preserving token positions~\citep{kvquant,KIVI}.


PAIR instead lets a smaller Reader prefill the full context and retain its KV cache.
The larger Reasoner does not prefill the full context, while the Reader retains all context positions and continues to contribute predictions throughout decoding.

\paragraph{Model collaboration.}

Existing model-collaboration methods are not designed to improve memory and compute efficiency. 
Speculative decoding aims to accelerate generation using a smaller draft model, but the larger verifier still processes the context and maintains its KV cache~\citep{specdec,specsampling}.
Minions aims to reduce cloud-model cost by decomposing long-context tasks between local and remote models and exchanging intermediate text~\citep{minions,cot}.
Request-routing methods like RouteLLM optimize the cost--quality trade-off by selecting a stronger or weaker model for each request, but do not reduce the inference cost of the stronger model when it is selected~\citep{routellm}.
DExperts, proxy-tuning, and contrastive or context-aware decoding combine predictions from multiple models for generation control, adaptation, or improved prediction quality~\citep{dexperts,proxytuning,contrastivedecoding,cad}.

CoGenesis~\citep{cogenesis} is the closest architectural precedent to PAIR.
It also combines a context-aware smaller model with a larger model that does not directly access the context, including token-level prediction fusion. However, CoGenesis focuses on private-context integration between local and cloud models rather than inference efficiency.

PAIR instead uses asymmetric context access to obtain a stronger accuracy--efficiency trade-off for long-context inference, reducing KV memory and computation while preserving answer accuracy.

\section{Motivation: how and why reader-reasoner fusion works}
\label{sec:motivation}
\subsection{Example: how combining reader and reasoner can help}
\label{sec:branch_motivation}

\vspace{-10pt}
\begin{figure*}[!ht]
\centering

\begin{subfigure}[t]{0.59\textwidth}
    \centering
    \includegraphics[width=\linewidth]{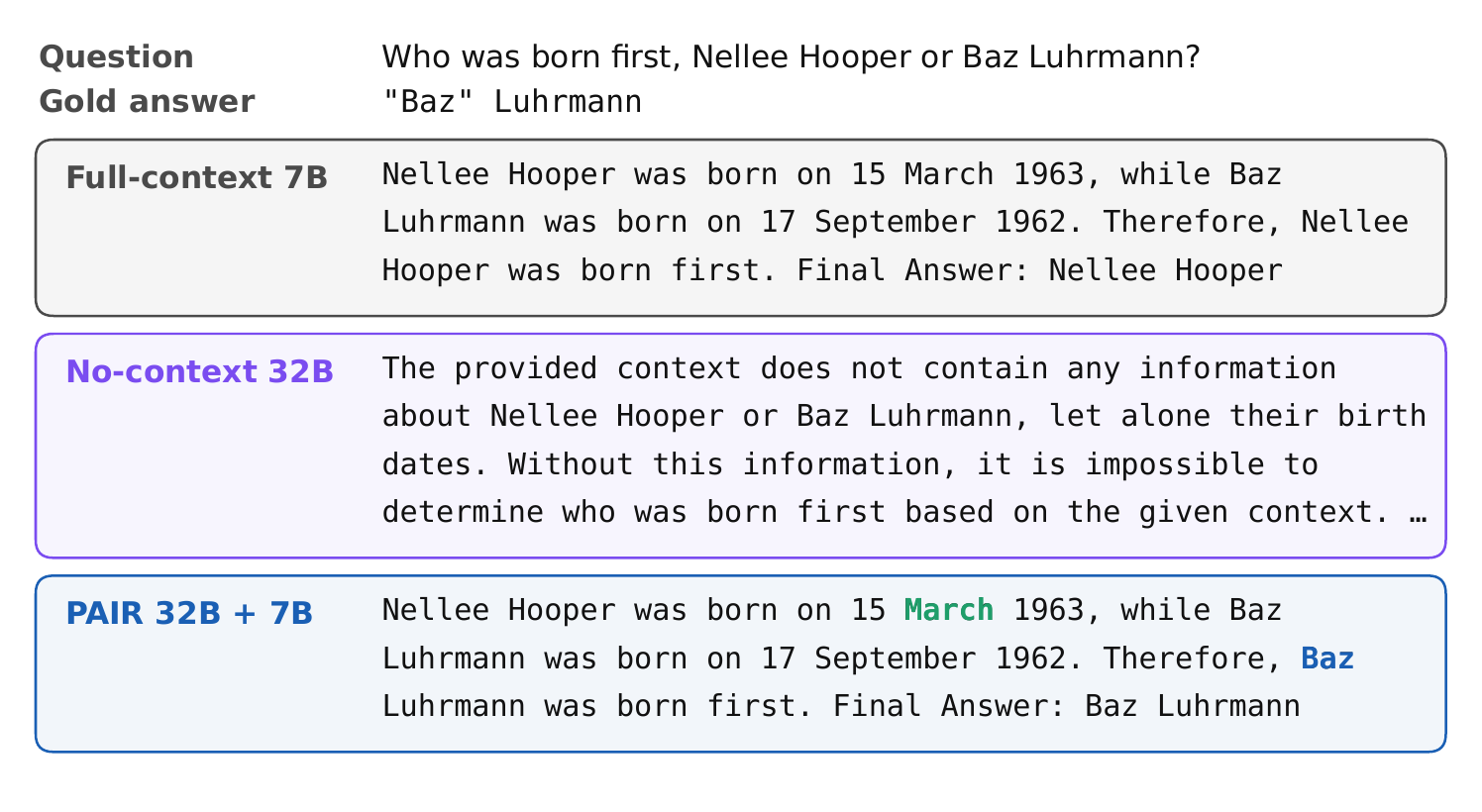}
    \vspace{-15pt}
    \caption{Generated answers.}
    \label{fig:motivating_example_answers}
\end{subfigure}
\hfill
\begin{subfigure}[t]{0.4\textwidth}
    \centering
    \includegraphics[width=0.9\linewidth]{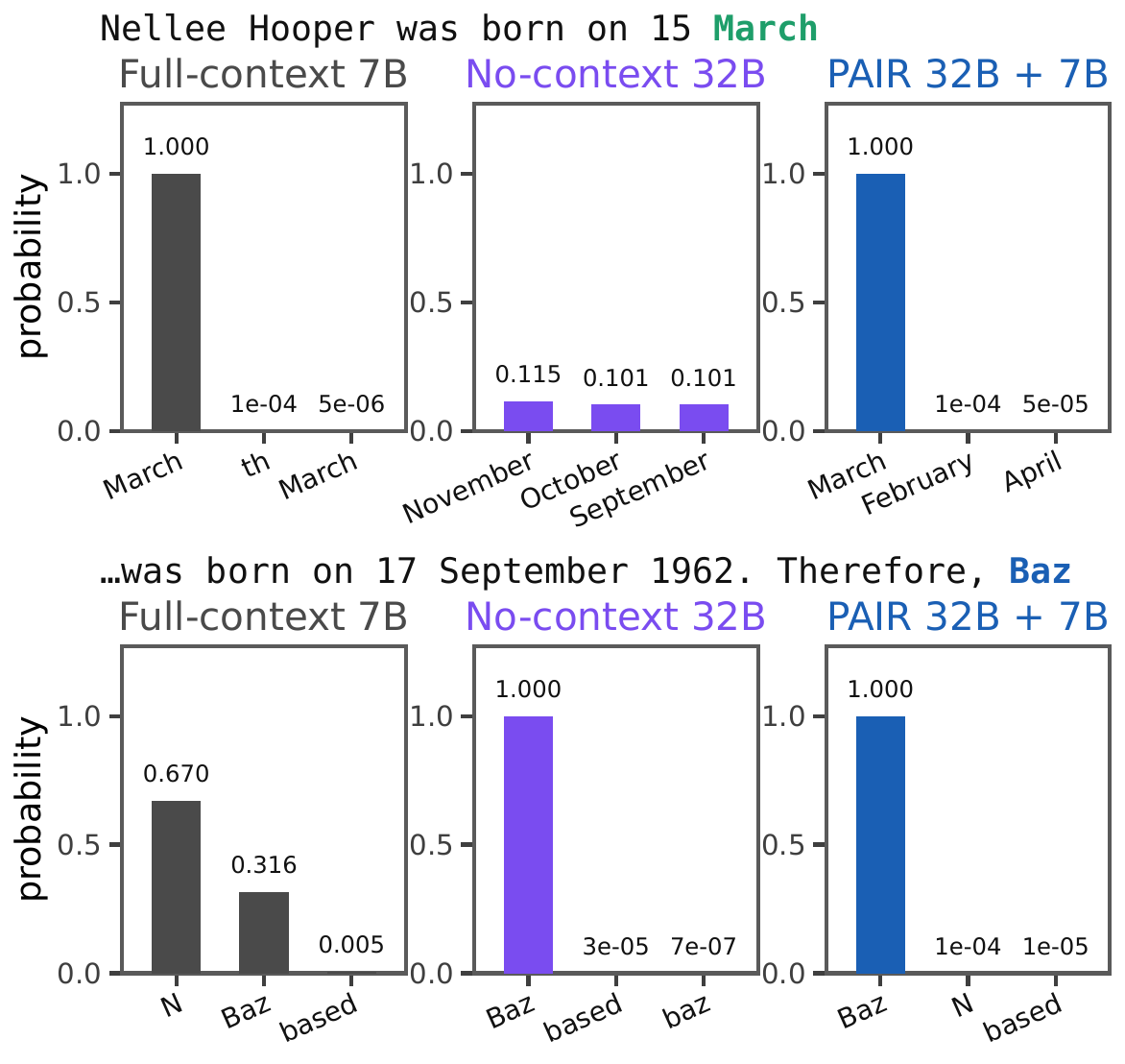}
    \vspace{-8pt}
    \caption{Next-token probabilities on the same PAIR-generated sequence.}
    \label{fig:motivating_example_topk}
\end{subfigure}
\vspace{-3pt}
\caption{
A HotpotQA example motivating PAIR. The example uses Qwen2.5 7B and 32B as the Reader and Reasoner, respectively.
(a) The small full-context model extracts both birth dates but reasons over them incorrectly, while the large no-context model lacks the relevant evidence and cannot determine the answer. PAIR correctly combines evidence extraction and answer reasoning to produce the answer.
(b) Teacher-forced next-token distributions along the PAIR-generated sequence. The top row evaluates an evidence-extraction position whose continuation depends on context-specific information, while the bottom row evaluates an answer-reasoning position after the relevant evidence has appeared in the generated sequence.
}
\vspace{-10pt}
\label{fig:motivating_example}
\end{figure*}

Processing the context with a smaller model is efficient, but the smaller model may still fail after finding the evidence needed to answer. \Cref{fig:motivating_example} shows one such HotpotQA example. The question asks whether Nellee Hooper or Baz Luhrmann was born first.
In panel (a), The small full-context model correctly states that Nellee Hooper was born on 15 March 1963 and Baz Luhrmann on 17 September 1962. The required evidence is already present in its generated sequence, but the model incorrectly concludes that Nellee Hooper was born first. In contrast, the large no-context model cannot determine the answer because it does not have access to the relevant evidence. PAIR combines these complementary capabilities and correctly answers that Baz Luhrmann was born first.

Panel (b) uses teacher forcing on the PAIR-generated sequence so that all three models are compared at the same decoding positions. The two positions expose different advantages of the Reader and Reasoner.
At the earlier evidence-extraction position, the full-context 7B model and PAIR assign all probability to the correct month ``March,'' while the no-context 32B model spreads its probability across several months.
At the later answer-reasoning position, the pattern reverses.
The full-context 7B model assigns higher probability to the incorrect continuation beginning with ``Nellee'' than to ``Baz,'' while the no-context 32B model and PAIR assign essentially all probability to ``Baz.''
These two positions show that the model with the stronger prediction can change during generation, with the full-context model providing context-specific evidence early and the larger no-context model providing stronger reasoning once that evidence appears in the generated sequence.

\subsection{Reader and Reasoner Provide Complementary Predictive Information}
\label{sec:complementarity}

In PAIR, the Reader and Reasoner have access to different information: the Reader has direct access to evidence in the full context, while the larger Reasoner can provide better predictions at reasoning-intensive positions. PAIR can benefit if the two models provide useful predictions at different positions. We quantify this complementarity using KL divergence to a full-context reference model, which we call the \emph{Teacher} ($T$). 
Under teacher forcing, all models predict the next token from the same Teacher-generated prefix. At decoding position $t$, let $p_{T,t}$, $p_{\mathrm{read},t}$, and $p_{\mathrm{reason},t}$ denote the resulting next-token distributions of the Teacher, Reader, and Reasoner. We measure the prediction error of model $m\in\{\mathrm{read},\mathrm{reason}\}$ at position $t$ by its KL divergence from the Teacher, 
$D_{m,t}=\mathrm{KL}(p_{T,t}\|p_{m,t})$,
and define its average as $D_m=\mathbb{E}_t[D_{m,t}]$. 
We also define an oracle that selects the model closer to the Teacher at each position as $D_{\mathrm{oracle}}=\mathbb{E}_t[\min(D_{\mathrm{read},t},D_{\mathrm{reason},t})]$.


For the 7B Reader and 32B Reasoner pair, the Reader has an average KL divergence of $D_{\mathrm{read}}=0.2885$, compared with $D_{\mathrm{reason}}=0.6869$ for the Reasoner. The token-wise oracle reaches $D_{\mathrm{oracle}}=0.1861$, reducing divergence by $35\%$ relative to the Reader. This gap shows that the Reasoner provides complementary predictive information at positions where it has lower KL divergence than the Reader. PAIR's fused prediction reaches $D_{\mathrm{PAIR}}=0.2338$, lower than either model alone, showing that weighted fusion recovers a substantial part of this complementary information without explicit routing. Evaluation details and the model-size sweep are provided in \Cref{app:complementarity}.

\section{Method: reader-reasoner fusion and role-specific tuning} 
\label{sec:method}
We first describe Reader--Reasoner inference and logit fusion, then present role-specific adaptation, and finally explain how PAIR balances the Reader and Reasoner predictions.
\subsection{Reader-reasoner inference and fusion at the logit level}

\begin{figure}[!t]
\centering
\includegraphics[width=0.9 \textwidth, trim={0 8 0 8}, clip]{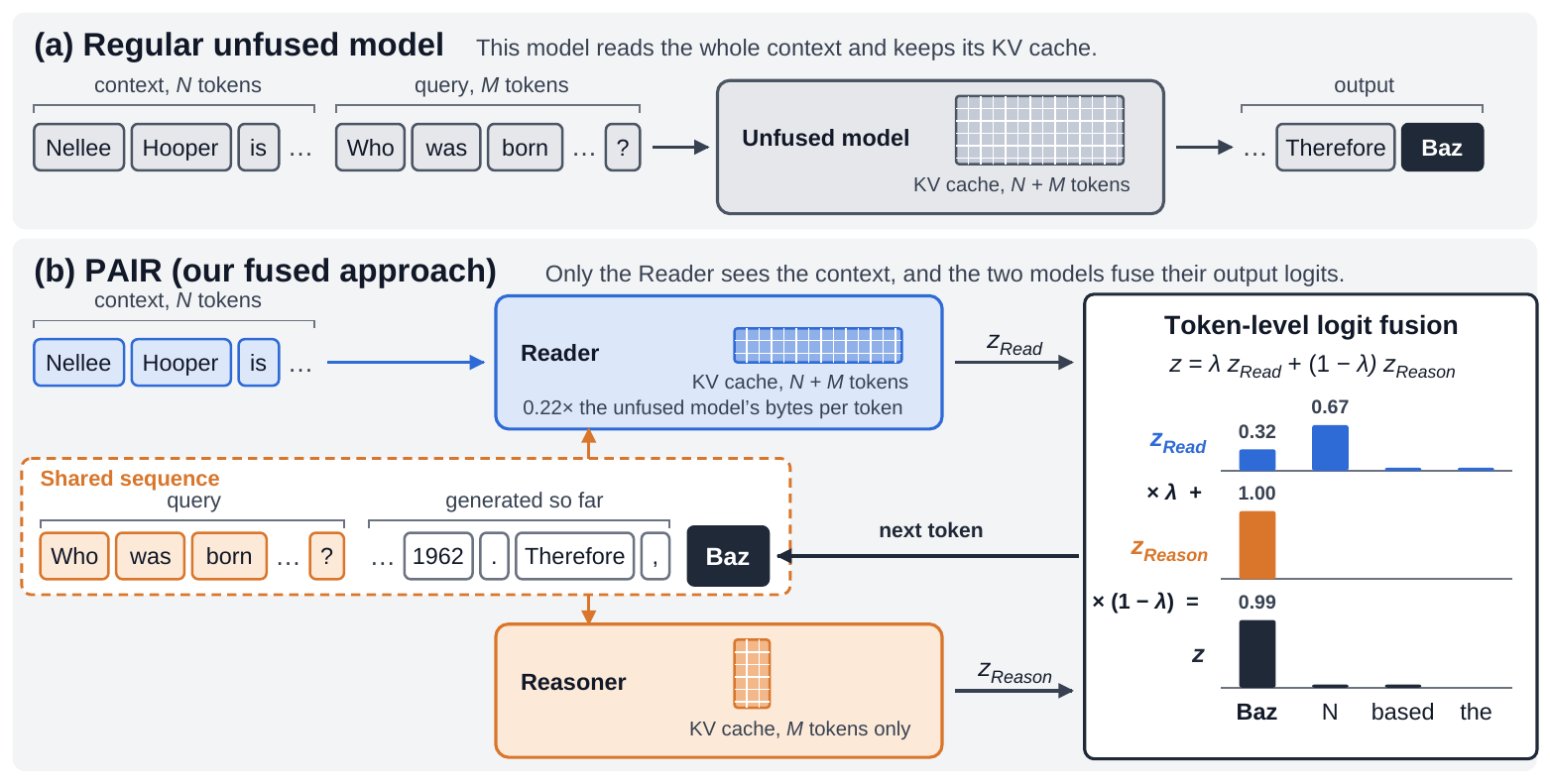}
\vspace{-5pt}
\caption{
Overview of PAIR.
The Reader prefills the full context and retains its KV cache, while the Reasoner does not attend to the full context.
At each decoding step, the Reader and Reasoner produce next-token predictions from the generated sequence, PAIR combines their logits, and the resulting token is appended to the sequence.
}
\vspace{-10pt}
\label{fig:overview}
\end{figure}

PAIR separates context reading for evidence extraction from reasoning over the extracted evidence to produce an answer, enabling memory-efficient and high-throughput inference.
Specifically, we let a small Reader attend to the full context to extract relevant evidence, while a larger Reasoner reasons over the extracted evidence without attending to the raw full context. PAIR combines their predictions throughout decoding rather than explicitly switching between the two roles.

As illustrated in \Cref{fig:overview}, let $x$ denote the full context, $q$ the question, and $y_{<t}$ the generated sequence before decoding step $t$.
The Reader conditions on $(x,q,y_{<t})$, while the Reasoner conditions on $(q,y_{<t})$ without attending to $x$.
They produce logits $z_{\mathrm{read},t}$ and $z_{\mathrm{reason},t}$, respectively, which we combine before sampling using a fixed fusion weight $\lambda\in[0,1]$,
\[
z_t
=
\lambda z_{\mathrm{read},t}
+
(1-\lambda) z_{\mathrm{reason},t},
\qquad
p(y_t \mid x,q,y_{<t})
=
\operatorname{softmax}(z_t).
\]
A single token $y_t$ is sampled from the fused distribution and appended to the generated sequence.
The updated sequence is then available to both models at the next decoding step.
The fusion weight $\lambda$ balances the contributions of the Reader and Reasoner. \Cref{sec:fusion_weight} analyzes this relationship and the empirical choice of $\lambda$.




\subsection{Role-specific adaptation through correlated fine-tuning}

The Reader and Reasoner have different inputs and responsibilities in PAIR.
The Reader must identify useful evidence from the full context and expose that information through the generated sequence.
The Reasoner must use the question and generated sequence to improve answer reasoning without directly accessing the context.
Off-the-shelf pretrained models are not trained specifically for these asymmetric roles, so we adapt the Reader and Reasoner separately with LoRA~\citep{lora}.
We adapt the Reader first because the Reasoner is subsequently trained with fusion enabled using the adapted Reader's predictions.

We first fine-tune the Reader in a standalone full-context setting.
At each step, the Reader predicts $p_{\mathrm{read}}(y_t \mid x, q, y_{<t})$ using the Teacher-generated sequence as the training target.
This adapts the Reader to better identify relevant evidence from the context.

We then fine-tune the Reasoner with fusion enabled.
The fine-tuned Reader is frozen at this stage, and the Reasoner conditions only on $(q,y_{<t})$ as during PAIR inference.
We form the fused prediction $p_{\mathrm{fuse},t}=\operatorname{softmax}(\lambda z_{\mathrm{read},t}+(1-\lambda)z_{\mathrm{reason},t})$ and train it on the same Teacher-generated sequence while updating only the Reasoner.
This stage adapts the Reasoner to contribute predictions that improve the fused output without directly accessing the context.

The full-context 32B Teacher remains frozen throughout both stages and provides generated reasoning and final answers as training targets. We use no gold reasoning or gold answers, and no evaluation examples are included in training. Once the Reader and Reasoner are adapted, the same weights are used across all benchmarks without benchmark-specific training or adaptation. Further training details are provided in Appendix~\ref{app:training_data}.

\subsection{Theoretical analysis behind fusion weight selection}
\label{sec:fusion_weight}

The fusion weight $\lambda$ determines how much the Reader and Reasoner contribute to each next-token prediction.
We view their predictions as two noisy estimates of a stronger full-context prediction, where larger prediction error favors more weight on the other model.
Let $e_{\mathrm{Read}}$ and $e_{\mathrm{Reason}}$ denote the Reader and Reasoner logit errors relative to the Teacher $T$.
Under the assumption that the Reader and Reasoner errors are zero-mean and uncorrelated, minimizing the expected squared fused error gives
$\lambda^\star=\sigma_{\mathrm{Reason}}^2/(\sigma_{\mathrm{Read}}^2+\sigma_{\mathrm{Reason}}^2)$,
where $\sigma_m^2$ denotes the error second moment of model $m$.
Near the reference distribution, KL divergence is locally approximated by a quadratic form of the logit error,
$\mathrm{KL}(p_T\|p_m)\approx \frac{1}{2}e_m^\top F_Te_m$,
where $F_T=\operatorname{diag}(p_T)-p_Tp_T^\top$.
We therefore use $D_m=\mathbb{E}_t[\mathrm{KL}(p_{T,t}\|p_{m,t})]$ as a proxy for this second moment, giving $\hat{\lambda}=D_{\mathrm{Reason}}/(D_{\mathrm{Read}}+D_{\mathrm{Reason}})$.

For HotpotQA and LoCoMo, it gives $\hat{\lambda}=0.70$ and $0.83$, respectively.
The larger value for LoCoMo reflects the substantially larger error of its Reasoner relative to the Reader, suggesting stronger dependence on evidence in context.
We then compare these predictions with empirical sweeps over $\lambda\in\{0.70,0.75,0.80,0.85\}$.
The sweeps select $\lambda=0.70$ for HotpotQA and $\lambda=0.85$ for LoCoMo, closely matching the corresponding estimates.
We use $\lambda=0.70$ for HotpotQA, MuSiQue, and LooGLE and $\lambda=0.85$ for LoCoMo.
Full sweeps and derivations are provided in \Cref{app:fusion_weight}.

\section{Evaluation}
\label{sec:experiment}
We evaluate four empirical claims about our proposed approach (PAIR).

\textbf{Accuracy and efficiency.}
We compare PAIR's accuracy and efficiency with the full-context 32B model and token-reduction baselines.
PAIR retains 89--96\% of its F1 score while using only 22.7--25.2\% of its KV memory and achieves 1.54--1.90$\times$ higher end-to-end throughput. It also provides a better accuracy-efficiency trade-off than the evaluated token-reduction baselines. See \Cref{sec:main_efficiency}.

\textbf{Generalization.}
We test PAIR's generalizability across model families and Reader-Reasoner size ratios. 
PAIR consistently improves over the corresponding small model across Qwen2.5 32B+7B and 32B+3B, OLMo-3 32B+7B, and Gemma-3 12B+4B, while remaining substantially closer to the corresponding large reference model with full context.
The same pattern holds across different Reader--Reasoner size ratios. See \Cref{sec:model_generalization}.

\textbf{Scaling.}
We tested how the efficiency of PAIR varies with changes in batch size and context length. PAIR remains faster than the full-context large model at matched batch sizes and moves toward Reader-level throughput as batch size increases. As context length grows, PAIR remains close to the full-context large model in accuracy while its throughput advantage increases. See \Cref{sec:scalability}.

\textbf{Role-specific adaptation.}
We test whether adapting both models is necessary for consistent accuracy recovery.
Plain fusion and Reader-only adaptation provide inconsistent gains, while adapting both models achieves the best fusion result across all four benchmarks.
See \Cref{sec:finetuning_ablation}.
\subsection{Experiment Setup}
\label{sec:setup}

\paragraph{Benchmarks.}
We evaluate PAIR on four long-context QA benchmarks: HotpotQA~\citep{hotpotqa}, MuSiQue~\citep{DBLP:journals/tacl/TrivediBKS22}, LooGLE~\citep{DBLP:conf/acl/LiWZZ24}, and LoCoMo~\citep{DBLP:conf/acl/MaharanaLTBBF24}. HotpotQA and MuSiQue are single-question tasks, where one question is answered from a long context. LooGLE and LoCoMo contain multiple questions associated with the same source context, which allows us to evaluate accumulated context use across turns.

For HotpotQA and MuSiQue, we expand each example to a 40-document context by retaining all original documents and adding BM25-retrieved distractors, as in retrieval-augmented pipelines~\citep{rag,sieve}.
For LooGLE and LoCoMo, we use an accumulated setting in which questions sharing the same context are asked sequentially. The context is processed once, and the Reader retains its KV cache across turns. We report token-level F1 on all benchmarks.
Detailed dataset construction and retrieval procedures are provided in \Cref{app:benchmarks}.

\paragraph{Generation and accumulated evaluation.}
Except for the model-generalization experiments in \Cref{sec:model_generalization}, all experiments use Qwen2.5 7B as the Reader and Qwen2.5 32B as the Reasoner.
Each response contains intermediate reasoning followed by a \texttt{Final Answer:} marker, and only the text after this marker is scored. The reasoning tokens provide decoding steps over which the Reader and Reasoner can combine their predictions.

HotpotQA and MuSiQue are evaluated one question at a time, with no state carried between examples. For LooGLE and LoCoMo, questions sharing the same context are evaluated sequentially. The Reader processes the full context only on the first turn and retains its KV cache across subsequent turns. After each response, the previous question, generated reasoning, and reference final answer are retained for the next turn. We replace the generated final answer with the reference answer to prevent answer errors from propagating through the history. The Reasoner is reset after every turn and begins each new question with only the instruction and current question, without the context or previous-turn history. During generation, the Reader and Reasoner receive the same generated sequence. Exact prompt, answer extraction, and history construction are in \Cref{app:generation_protocol}.

\paragraph{Baselines.}
We compare PAIR with the full-context 32B and 7B models and with SnapKV~\citep{snapkv}, ExpectedAttention~\citep{expectedattention}, and SpecPrefill~\citep{specprefill} as representative context reduction baselines. We also evaluate full-model KV quantization as an orthogonal memory-reduction approach that can be combined with PAIR. Their configurations are in \Cref{app:baselines}, with the quantization results in \Cref{app:quantization}.



\subsection{Preserving Accuracy with Less KV Memory and Computation}
\label{sec:main_efficiency}

We examine whether PAIR can preserve most of the full-context 32B reference model's accuracy while reducing KV memory and increasing throughput.

We compare PAIR with the full-context 32B model, the full-context 7B Reader, and three token-reduction baselines, SnapKV, ExpectedAttention, and SpecPrefill. The full-context models and PAIR each use one configuration per benchmark, while the token-reduction baselines are evaluated across multiple retention budgets. We measure answer F1, KV memory size normalized by the full-context 32B model, and time per completed answer, computed as the inverse of aggregate throughput. Timing includes both context prefill and decoding. Throughput is measured at the largest tested feasible batch for each method and configuration on the same NVIDIA GH200 node. KV-memory accounting, batch selection, and exact operating points in \Cref{app:efficiency}.

\begin{figure*}[!htbp]
\centering
\includegraphics[width=\textwidth, trim={0 8 0 8}, clip]{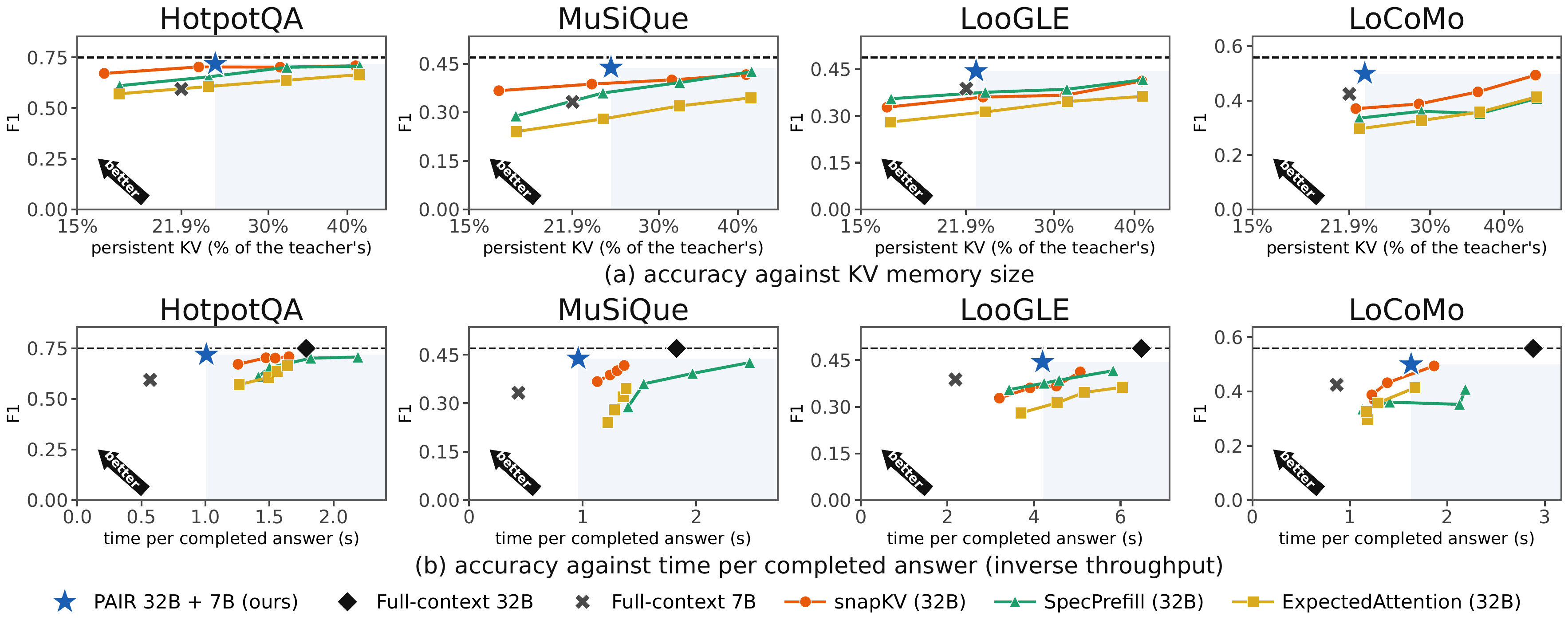}
\vspace{-15pt}
\caption{
Accuracy--efficiency trade-offs across the four benchmarks.
(a) F1 against KV memory normalized by the full-context 32B model.
(b) F1 against time per completed answer, the inverse of aggregate throughput, where lower values indicate higher throughput. 
Arrows indicate the direction of improvement.
Regions marked blue \pairshade\ contain operating points that are worse than PAIR on both axes.
Horizontal dashed lines indicate the full-context 32B model's accuracy.
}
\vspace{-5pt}
\label{fig:main_efficiency}
\end{figure*}

\Cref{fig:main_efficiency}(a) shows that PAIR preserves most of the full-context 32B accuracy with substantially less KV memory. On HotpotQA, PAIR reaches 0.7178 F1 versus 0.7492 while using 24.8\% of the full-context 32B model's KV memory. Across the four benchmarks, PAIR retains 89--96\% of its F1 with only 22.7--25.2\% of its KV memory. PAIR also achieves higher F1 than every evaluated SnapKV, ExpectedAttention, and SpecPrefill operating point, including those that use more KV memory.

\Cref{fig:main_efficiency}(b) shows the corresponding throughput results. PAIR supports a larger maximum feasible batch size, $B_{\max}$, than the full-context 32B model and achieves 1.54--1.90$\times$ higher throughput across the four benchmarks, with each method evaluated at its respective $B_{\max}$. It also achieves higher F1 and higher throughput than the highest-accuracy operating point of every evaluated token-reduction baseline. \Cref{sec:scalability} separates the effect of larger batch capacity from execution cost and studies how both change with batch size and context length.

PAIR provides both accuracy and throughput benefits across workload types. The larger gain differs by setting. In single-question settings, assigning full-context prefill to the smaller Reader yields substantial throughput gains. In accumulated settings, retaining the full context in the Reader preserves evidence across turns, leading to a clear accuracy advantage over token-reduction baselines.

PAIR can also reduce per-request latency when the Reader and Reasoner are placed on separate GPUs. Because Reader--Reasoner interaction occurs only through logit fusion, separate GPUs enable parallel decoding. Our two-GPU evaluation shows lower latency without reducing aggregate throughput under the same hardware budget. Detailed results are provided in \Cref{app:two_gpu}.

\subsection{Accuracy Preservation Generalizes Across Model Families}
\label{sec:model_generalization}
We ask whether PAIR preserves accuracy across different model families and Reader--Reasoner size ratios, rather than only for the Qwen2.5 32B Reasoner and 7B Reader used in our main experiments.

We evaluate four model pairs spanning three families and multiple Reader sizes, including Qwen2.5 32B+7B, Qwen2.5 32B+3B, OLMo-3 32B+7B, and Gemma-3 12B+4B, using the same HotpotQA setting as in \Cref{sec:main_efficiency}. In each pair, the larger model serves as the Reasoner and the smaller model as the Reader. Each family uses its own Teacher-generated targets and trained adapters. Additional details and exact scores are provided in \Cref{app:model_generalization}.

\begin{figure*}[t]
\centering
\includegraphics[width=\textwidth]{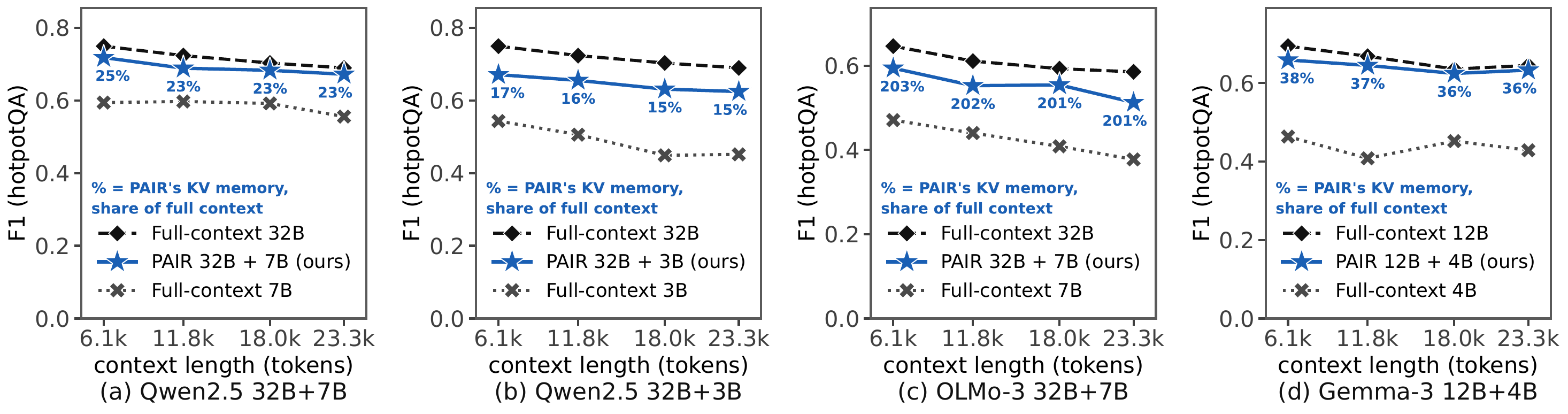}
\vspace{-15pt}
\caption{
Accuracy generalization across model families and Reader sizes on HotpotQA.
Each plot reports F1 for the full-context large model, PAIR, and the corresponding Reader at varying context lengths.
Blue percentages below the PAIR curve report its KV memory relative to the corresponding full-context large model.
For OLMo-3, the ratios exceed 100\% because the 7B model uses MHA with 32 KV heads, while the 32B model uses GQA with 8 KV heads~\citep{gqa}.
}
\vspace{-10pt}
\label{fig:model_generalization}
\end{figure*}

\Cref{fig:model_generalization} shows that this behavior persists across all four model pairs. As context length increases, PAIR consistently improves over the corresponding Reader-only baseline while remaining substantially closer to the full-context large model. This pattern holds when the Reader is reduced to Qwen2.5 3B and when both models are replaced with OLMo-3 or Gemma-3. PAIR's accuracy advantage therefore does not depend on a single model family or a particular Reader size.
For OLMo-3, the KV-memory efficiency claim does not apply because the 7B model uses MHA while the 32B model uses GQA. We include this pair only to evaluate accuracy generalization.

\subsection{PAIR Moves Toward Reader-Level Efficiency as Workloads Scale}
\label{sec:scalability}

The main throughput results use the largest feasible batch for each method on the same 95-GiB GPU. This comparison reflects both execution cost and the batch capacity available under that memory limit. We therefore measure throughput at matched batch sizes to separate these effects. We also vary context length to test whether PAIR continues to preserve accuracy while maintaining its throughput advantage in longer inputs. See \Cref{app:scalability} for experimental details and complete operating points.

\begin{figure*}[!ht]
\centering
\includegraphics[width=0.95\textwidth]{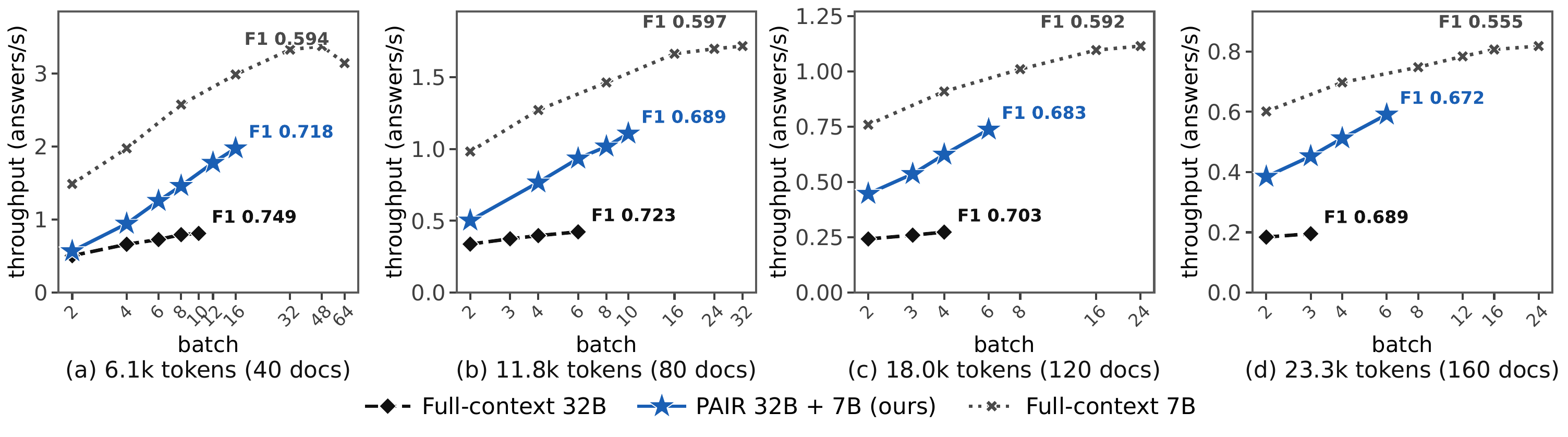}
\vspace{-10pt}
\caption{
End-to-end throughput across batch sizes at four context lengths on the same 95-GiB GPU used for the main throughput experiments.
F1 annotations report the accuracy of each method at the corresponding context length.
PAIR remains close to the full-context 32B model in F1 while moving toward the 7B Reader in throughput as batch size increases.
}
\vspace{-10pt}
\label{fig:batch_scaling}
\end{figure*}

\paragraph{Larger batches move PAIR toward Reader throughput.}
We ask whether PAIR remains faster than the full-context 32B model at matched batch sizes and moves toward Reader-level throughput as batch size increases.

To separate execution cost from the different batch capacities under a fixed memory budget, we sweep matched batch sizes for the full-context 32B model, the 7B Reader, and PAIR at four context lengths on the same 95-GiB GPU.

\Cref{fig:batch_scaling} shows that PAIR is faster than the full-context 32B model at every matched batch size and moves closer to the Reader as batch size increases.
At 6.1k tokens, the Reader-to-PAIR throughput ratio decreases from 2.62$\times$ at batch 2 to 1.51$\times$ at batch 16, with the same trend across the other context lengths.
Both PAIR and the Reader process the full context with the 7B model, while PAIR additionally runs the 32B Reasoner on the question and generated sequence.
The relative cost of the Reasoner therefore decreases at larger batches.

\paragraph{PAIR retains its advantage as context length grows.}
We ask whether PAIR continues to preserve accuracy while maintaining its throughput advantage as the context becomes longer.

We evaluate the same HotpotQA questions with varying context lengths from 6.1k to 45.7k tokens.

\begin{figure*}[!ht]
\centering
\includegraphics[width=\textwidth]{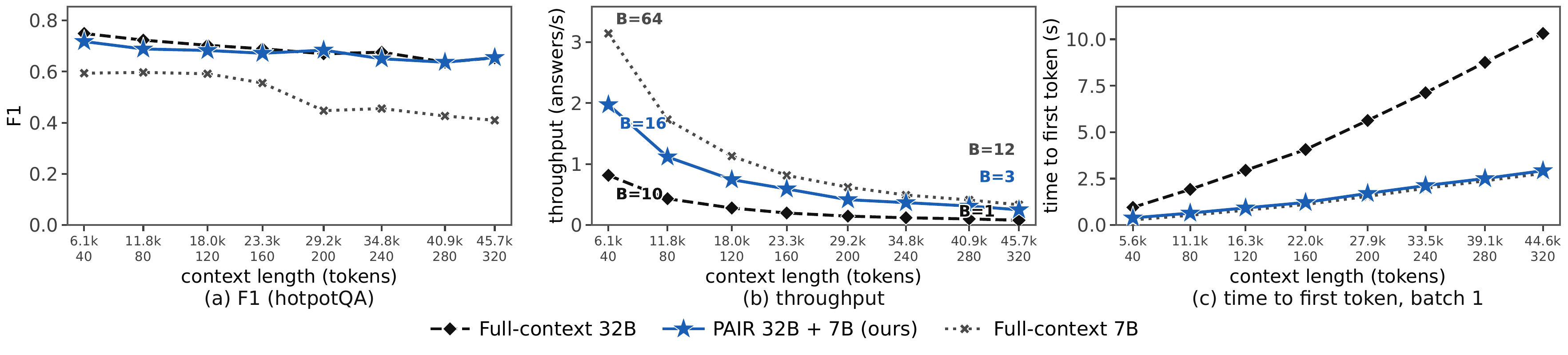}
\vspace{-20pt}
\caption{
Context-length scaling on HotpotQA.
The left plot reports F1, the center plot reports end-to-end throughput at the largest tested feasible batch size, and the right plot reports time to first token at batch size one.
PAIR remains close to the full-context 32B model in accuracy while its throughput advantage increases with context length.
}
\vspace{-10pt}
\label{fig:context_scaling}
\end{figure*}

\Cref{fig:context_scaling}(a) shows that PAIR remains close to the full-context 32B model in accuracy across the full range. \Cref{fig:context_scaling}(b) shows that its throughput advantage increases from 2.42$\times$ at 6.1k tokens to 3.20$\times$ at 45.7k tokens. PAIR processes the growing context with the 7B Reader rather than the 32B Reasoner, so the Reasoner's relative cost decreases as context length grows.

\subsection{Adapting both the Reader and Reasoner is necessary.}
\label{sec:finetuning_ablation}
We test whether adapting both the Reader and Reasoner is necessary for PAIR to recover accuracy.

\begin{table}[!htbp]
\centering
\footnotesize
\setlength{\aboverulesep}{0pt}
\setlength{\belowrulesep}{0pt}
\renewcommand{\arraystretch}{1.0}
\caption{
Fine-tuning ablation on the four primary benchmarks.
Reader FT adapts only the Reader, while Reader + Reasoner FT
adapts both models and corresponds to PAIR.
Shaded columns show the full-context reference baselines;
bold indicates the best result among the fusion variants.
}
\label{tab:finetuning_ablation}
\vspace{-5pt}
\begin{tabular}{
    l
    >{\columncolor{gray!10}}c
    ccc
    >{\columncolor{gray!10}}c
}
\toprule
Benchmark & Full-context 7B & Plain fusion & Reader FT
& Reader + Reasoner FT & Full-context 32B \\
\midrule
HotpotQA & 0.5939 & 0.6720 & 0.6633 & \textbf{0.7178} & 0.7492 \\
MuSiQue  & 0.3323 & 0.3716 & 0.3939 & \textbf{0.4380} & 0.4695 \\
LooGLE   & 0.3878 & 0.3654 & 0.3658 & \textbf{0.4438} & 0.4876 \\
LoCoMo   & 0.4243 & 0.4255 & 0.4916 & \textbf{0.4992} & 0.5583 \\
\bottomrule
\end{tabular}
\vspace{-5pt}
\end{table}

We ablate the two adaptation stages in \Cref{tab:finetuning_ablation}.
Plain fusion uses the unadapted Reader and Reasoner, Reader FT adapts only the Reader, and Reader + Reasoner FT adapts both models.

Plain fusion and Reader-only adaptation provide inconsistent gains, while adapting both models achieves the best fusion result on all four benchmarks.
This shows that adapting the Reasoner under fused decoding is necessary for consistent accuracy recovery.


\section{Limitation}
Although its accuracy-preserving behavior generalizes across multiple model families and sizes, this generalization is empirical and not theoretically guaranteed. PAIR currently uses a fixed fusion weight for each setting, and learning or dynamically adapting this weight may further improve accuracy. Finally, our experiments cover contexts up to 45.7k tokens, and evaluating PAIR at substantially longer context lengths remains future work.

\section{Conclusion}
Answering from a given context with a large model is costly because the model must process and retain the full context. PAIR instead combines a smaller Reader that processes the context with a larger Reasoner that does not directly attend to it. PAIR retains 89--96\% of the full-context large model's F1 while using only 22.7--25.2\% of its KV memory and achieves a better accuracy--throughput tradeoff than the evaluated token-reduction baselines. This accuracy preservation generalizes across different model sizes and families. As batch size and context length increase, PAIR moves toward Reader-level efficiency while remaining close to the full-context large model's accuracy.

\section*{Acknowledgments}

This work was supported in part by the National Science Foundation under grants \#2440498 and \#2312561. This work also used computing resources provided through allocations from the Advanced Cyberinfrastructure Coordination Ecosystem: Services \& Support (ACCESS) program, which is supported by U.S. National Science Foundation grants \#2138259, \#2138286, \#2138307, \#2137603, and \#2138296.

\section*{AI Use Statement}

We used generative AI tools to assist with drafting and revising the manuscript, experiment implementation, and debugging. We also used a frozen full-context Teacher model to generate distillation targets for the Reader and Reasoner, as described in Section~\ref{sec:method}. All AI-assisted work was reviewed and verified by the authors, who take responsibility for the final text, code, results, and claims.

\section*{Ethics statement}

This work studies the accuracy and efficiency of long-context language model inference using existing question-answering benchmarks. PAIR inherits potential biases and factual errors from its constituent models and Teacher-generated training targets. Retaining the full context does not guarantee that the generated answer faithfully reflects the source.

\section*{Reproducibility statement}

Anonymous source code and experimental artifacts are available at \artifacturl. The main paper specifies the PAIR architecture, training procedure, evaluation metrics, and primary experimental setup. The appendix provides additional details on benchmark construction, training, baselines, efficiency measurements, scalability experiments, and ablations.

\bibliography{iclr2027_conference}
\bibliographystyle{iclr2027_conference}

\appendix
\section*{Appendix}

\section{Branch Complementarity Evaluation}
\label{app:complementarity}
In this appendix, we call each of the Reader and Reasoner paths in PAIR a \emph{branch}.

We evaluate branch complementarity on 240 HotpotQA examples under teacher forcing. The full-context 32B Teacher first generates one reference trajectory for each example via greedy decoding. The Reader, Reasoner, and fused configurations are then evaluated on the same stored trajectory, so all models are conditioned on identical prefixes at every position. We compute $\mathrm{KL}(p_T\|p_m)$ over the full vocabulary and report the mean in nats.

We evaluate the reasoning and answer span while excluding the \texttt{Final Answer} marker tokens and any tokens after the Teacher's first \texttt{<|endoftext|>}. For the headline analysis, we also exclude the Teacher's terminal \texttt{<|endoftext|>} token itself. This leaves 17,012 evaluated positions across the 240 examples. The terminal token is excluded because it primarily measures whether a model chooses to stop at exactly the same position as the Teacher. It has a disproportionate effect on the fused models while contributing little information about answer correctness.

\subsection{Position-wise Branch Divergence}
\label{app:example_kl}

\begin{figure}[ht]
\centering
\includegraphics[width=\columnwidth]{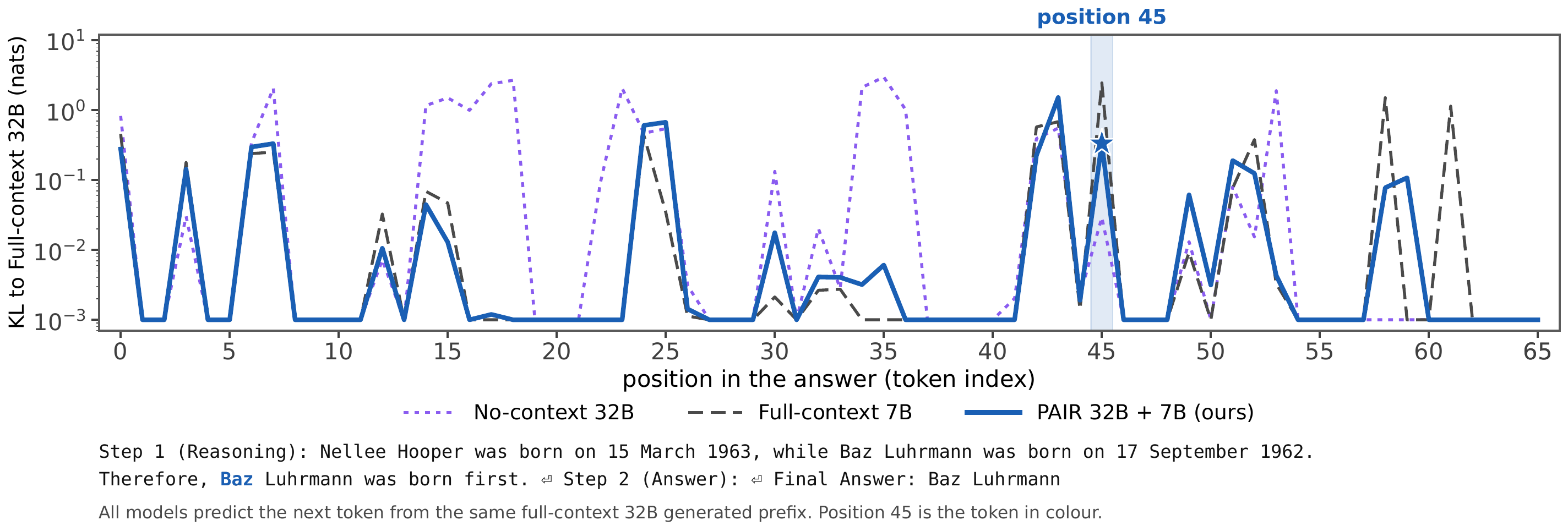}
\vspace{-4pt}
\caption{
Position-wise KL divergence to the full-context 32B Teacher for the motivating HotpotQA example.
The Reader, Reasoner, and PAIR are evaluated on the same Teacher-generated trajectory.
The highlighted position corresponds to the decisive token, ``Baz'' shown in \Cref{sec:branch_motivation}.
}
\label{fig:example_branch_kl}
\end{figure}

We additionally inspect the HotpotQA example used in \Cref{sec:branch_motivation}. Unlike the free-running outputs shown in the main text, this analysis evaluates the Reader, Reasoner, and PAIR on the same Teacher-generated trajectory. \Cref{fig:example_branch_kl} reports their KL divergence to the full-context 32B Teacher at each decoding position.

The branch divergences vary substantially across the sequence. At some positions the Reader is closer to the Teacher, while at others the Reasoner is closer. PAIR often remains closer to the lower-divergence branch when the two predictions differ. The highlighted position corresponds to the decisive token in the motivating example, where the generation continues toward ``Baz'' rather than the incorrect ``Nellee'' by the Reader. This example illustrates the position-dependent complementarity measured more broadly in the aggregate analysis below.

\subsection{Main 7B Reader and 32B Reasoner}

For the main configuration, we compute $D_{\mathrm{Read}}=\mathbb{E}_t[\mathrm{KL}(p_{T,t}\|p_{R,t})]$, $D_{\mathrm{Reason}}=\mathbb{E}_t[\mathrm{KL}(p_{T,t}\|p_{Q,t})]$, and $D_{\mathrm{oracle}}=\mathbb{E}_t[\min(D_{R,t},D_{Q,t})]$. The oracle selects the branch closer to the Teacher independently at each position. It requires access to the Teacher and is therefore used only as a reference for the predictive coverage available from the Reader and the Reasoner.

PAIR uses the deployed fusion weight $\lambda=0.70$ and both trained adapters. We also evaluate plain fusion using the same Reader and Reasoner with $\lambda=0.70$ but without fine-tuning.

\begin{table}[ht]
\centering
\small
\caption{
KL divergence to the full-context 32B Teacher on the main HotpotQA configuration.
Gap closed measures progress from the Reader toward the token-wise oracle.
}
\label{tab:branch_complementarity_main}
\begin{tabular}{lcc}
\toprule
Configuration & KL to Teacher & Gap closed \\
\midrule
Reader 7B & 0.2885 & -- \\
Reasoner 32B & 0.6869 & -- \\
Plain fusion & 0.2428 & 45\% \\
PAIR 32B + 7B (Ours) & \textbf{0.2338} & 53\% \\
Oracle & 0.1861 & 100\% \\
\bottomrule
\end{tabular}
\end{table}

The oracle reduces divergence by $35\%$ relative to the Reader, showing that the Reasoner is closer to the Teacher at positions where the Reader is not. Plain fusion already reduces divergence to 0.2428 and closes $45\%$ of the Reader-to-oracle gap. Role-specific fine-tuning further reduces divergence to 0.2338, allowing PAIR to close $53\%$ of the gap.

\subsection{Model-Size Sweep}

We also evaluate all 21 Qwen2.5 Reader–Reasoner pairs drawn from model sizes
\(\{0.5\mathrm{B},1.5\mathrm{B},\) \(3\mathrm{B},7\mathrm{B},14\mathrm{B},32\mathrm{B},72\mathrm{B}\}\)~\citep{qstore}, assigning the smaller model to the Reader in each pair. No fine-tuning or fusion is used in this sweep. For each pair, the Reader receives the document and the question, while the Reasoner receives the question without the document. The Teacher is the same model as the Reasoner with access to the document.

\begin{figure}[t]
\centering
\includegraphics[width=0.82\textwidth]{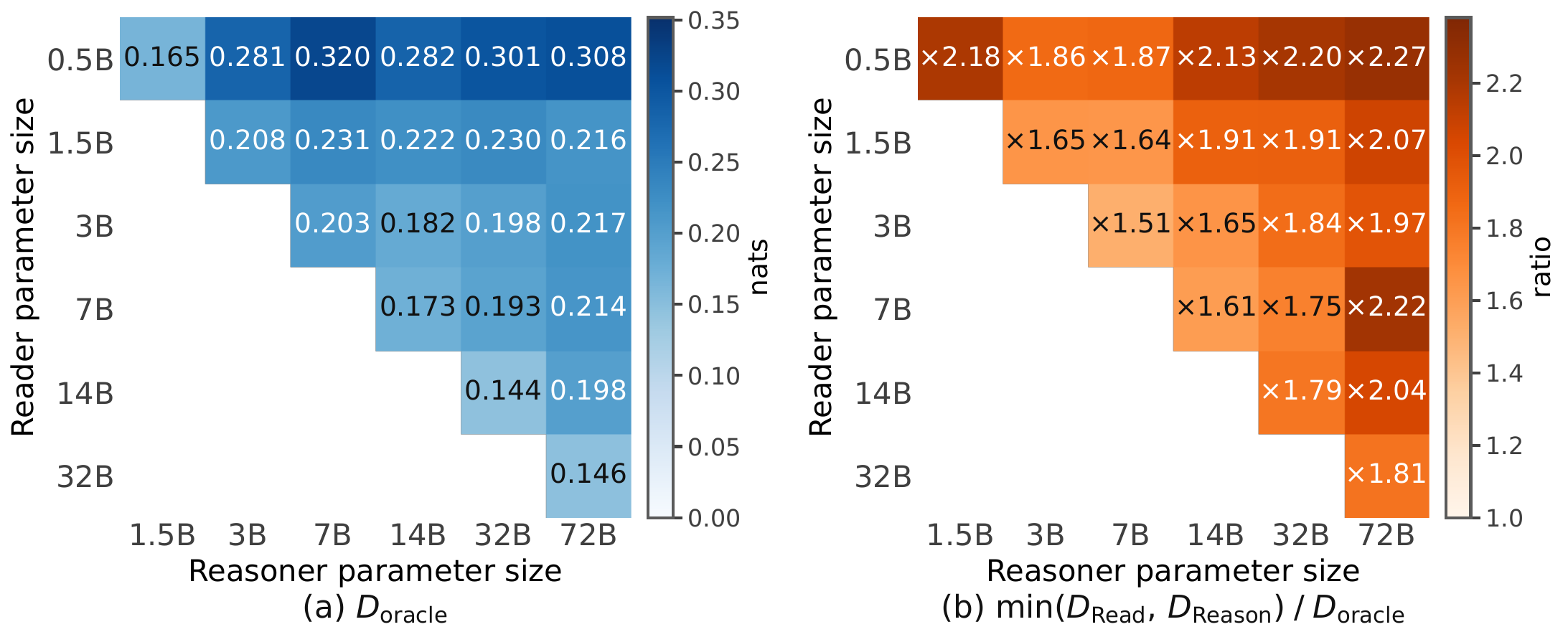}
\vspace{-4pt}
\caption{
Branch complementarity across Qwen2.5 model scales under teacher forcing.
Rows vary the Reader size and columns the Reasoner size.
The left plot reports token-wise oracle divergence $D_{\mathrm{oracle}}$.
The right plot reports $\min(D_{\mathrm{Read}},D_{\mathrm{Reason}})/D_{\mathrm{oracle}}$.
The model-size sweep includes the Teacher's terminal token.
}
\label{fig:branch_complementarity}
\end{figure}

\Cref{fig:branch_complementarity} reports $D_{\mathrm{oracle}}$ and the complementarity ratio $\min(D_{\mathrm{Read}},D_{\mathrm{Reason}})/D_{\mathrm{oracle}}$ across all pairs. This sweep uses the original branch-only evaluation convention and includes the Teacher's terminal \texttt{<|endoftext|>} token. Under this convention, the main 7B Reader and 32B Reasoner pair gives $(D_{\mathrm{Read}},D_{\mathrm{Reason}},D_{\mathrm{oracle}})=(0.337,0.689,0.193)$ and a complementarity ratio of $1.75$. The difference from the headline values above comes from the treatment of the terminal token.

Across all 21 pairs, the complementarity ratio ranges from $1.51$ to $2.27$, corresponding to reductions of approximately $34\%$ to $56\%$ relative to the better standalone branch. The effect therefore persists across a wide range of Reader and Reasoner capacities.

We estimate uncertainty by bootstrap resampling the 240 examples 2,000 times. The standard error of the complementarity ratio ranges from 0.02 to 0.05 across model pairs.

\section{Training Data and Adaptation Details}
\label{app:training_data}

\subsection{Training Corpora}

The Reader adapter is trained on our S-solo corpus. The corpus contains 4,819 synthetic hard-retrieval examples and 2,000 real-data examples drawn from 2Wiki~\citep{2wiki}, SciFact~\citep{scifact}, and LongBench~\citep{longbench}. S-solo is used to specialize the Reader for long-context retrieval and reasoning.

The Reasoner fusion adapter is trained using our L-fusion corpus. L-fusion contains 4,312 synthetic binding examples and 2,000 real-data examples drawn from 2Wiki, SciFact, and LongBench.

For both corpora, supervision consists of the full-context Teacher's generated reasoning followed by its final answer. Teacher generations are used as distillation targets without filtering them according to whether the generated final answer agrees with the dataset's gold answer. The objective is therefore to transfer the Teacher's output behavior rather than to construct a separately gold-corrected training set. No samples from the evaluation set were used during the fine-tuning process.

\subsection{Synthetic Binding Tasks}

The synthetic component of L-fusion contains 4,312 examples, each constructed from a distinct generated document. The documents consist of streams of schema-identical records containing recurring attribute slots such as dates, people, locations, and versions. Questions request one attribute of one event, but the target event is not identified directly by its record identifier.

Solving an example therefore requires two steps. The model must first bind the question to the correct record among many structurally similar distractors. It must then resolve an indirect reference within that record to obtain the requested value.

The 4,312 examples use 737 unique question strings. Question wording repeats across documents because the generator draws from a finite set of templates, while the underlying documents and record values differ.

We vary three aspects of the synthetic task. First, the same record stream is rendered in seven surface formats: memo, changelog, notebook, ticket, minutes, specification, and transcript. These variants are intended to reduce dependence on a particular serialization format. Second, the task spans seven reference types: relative time, speaker, locative reference, unit reference, ordinal reference, ellipsis, and scope. Each reference type contributes 616 examples. Third, examples are distributed across four content domains: releases, incidents, holdings, and studies, with 1,078 examples per domain.

The 2,000 real-data examples in L-fusion contain 1,773 distinct documents, with up to 10 questions associated with the same document.

The task design was motivated by errors observed during development on long-context and accumulated-context workloads, including failures in which the model retrieved the relevant region but failed to bind an indirect reference to the appropriate event. No evaluation examples are copied into the training corpus.

\subsection{Binding-Task Example}

Consider a synthetic example with
\texttt{container=ticket},
\texttt{domain=releases}, and
\texttt{reftype=reltime}.
The document contains 24 schema-identical ticket records, several of which describe different software releases. The question asks:

\begin{quote}
On what date did the indexer release take place?
\end{quote}

The relevant record is:

\begin{quote}
\texttt{[TICKET-1019] opened 30 October 2022 ... I shipped the indexer release the week before.}
\end{quote}

Other records contain distractors such as a cache release, parser release, and gateway release. Answering the question requires first binding ``indexer release'' to \texttt{TICKET-1019}, then interpreting ``the week before'' relative to that ticket's opening date.

Teacher supervision is retained exactly as generated. Consequently, a Teacher mistake is not corrected by using the synthetic gold label before distillation. In this particular example, the Teacher incorrectly resolved the temporal anchor and produced an incorrect date; the generated response remains the training target. This is consistent with our objective of distilling the Teacher's behavior rather than training directly on synthetic gold answers.

\section{Fusion Weight Analysis Details}
\label{app:fusion_weight}

\subsection{Fusion Weight Sweep}

We sweep fusion weight $\lambda\in\{0.70,0.75,0.80,0.85\}$ under matched training and inference settings.
Because changing $\lambda$ requires separately adapting the Reasoner under the corresponding fusion setting, we perform the full sweep on HotpotQA and LoCoMo.
MuSiQue and LooGLE use the common $\lambda=0.70$ setting.

\begin{table}[ht]
\centering
\small
\caption{
F1 under matched fusion-weight sweeps.
}
\label{tab:lambda_sweep}
\begin{tabular}{lcccc}
\toprule
Benchmark &
$\lambda=0.70$ &
$\lambda=0.75$ &
$\lambda=0.80$ &
$\lambda=0.85$ \\
\midrule
LoCoMo & 0.461 & 0.486 & 0.495 & \textbf{0.499} \\
HotpotQA & \textbf{0.718} & 0.710 & 0.712 & 0.710 \\
\bottomrule
\end{tabular}
\end{table}

LoCoMo improves as the mixture becomes more Reader-heavy, while HotpotQA does not benefit from increasing $\lambda$ above $0.70$.

\subsection{Minimum-Variance Derivation}

Let the two branch estimates be

\[
\hat{z}_{\mathrm{Read}} = z^\star + e_{\mathrm{Read}},
\qquad
\hat{z}_{\mathrm{Reason}} = z^\star + e_{\mathrm{Reason}},
\]

and let their linear combination be

\[
\hat{z}
=
\lambda \hat{z}_{\mathrm{Read}}
+
(1-\lambda)\hat{z}_{\mathrm{Reason}}.
\]

Under the assumption of zero-mean and uncorrelated errors with variances
$\sigma_{\mathrm{Read}}^2$ and $\sigma_{\mathrm{Reason}}^2$,

\[
\mathbb{E}\left[\|\hat{z}-z^\star\|^2\right]
=
\lambda^2\sigma_{\mathrm{Read}}^2
+
(1-\lambda)^2\sigma_{\mathrm{Reason}}^2.
\]

Differentiating with respect to $\lambda$ gives

\[
2\lambda\sigma_{\mathrm{Read}}^2
-
2(1-\lambda)\sigma_{\mathrm{Reason}}^2
=
0,
\]

and therefore

\[
\lambda^\star
=
\frac{\sigma_{\mathrm{Reason}}^2}
{\sigma_{\mathrm{Read}}^2+\sigma_{\mathrm{Reason}}^2}.
\]

The independence assumption is used only for this interpretation. With correlated branch errors, the optimum additionally depends on their covariance.

\subsection{Local KL Interpretation}

Let $z_T$ denote the full-context Teacher logits, and let
\[
z_m = z_T + e_m,
\qquad
p_T = \operatorname{softmax}(z_T),
\qquad
p_m = \operatorname{softmax}(z_T + e_m).
\]

Consider the Teacher-relative divergence as a function of the logit
perturbation,
\[
f(e_m)
=
\mathrm{KL}
\left(
p_T
\;\|\;
\operatorname{softmax}(z_T+e_m)
\right).
\]
A second-order Taylor expansion around $e_m=0$ gives
\[
f(e_m)
=
f(0)
+
\nabla f(0)^\top e_m
+
\frac{1}{2}
e_m^\top
\nabla^2 f(0)
e_m
+
O(\|e_m\|^3).
\]

At $e_m=0$, the two distributions are identical, so
$f(0)=\mathrm{KL}(p_T\|p_T)=0$. This point is also the minimum of the
divergence, giving $\nabla f(0)=0$. For the softmax distribution,
\[
\nabla^2 f(0)
=
\operatorname{diag}(p_T)-p_Tp_T^\top
\equiv F_T,
\]
which is the Fisher information matrix of the categorical distribution
induced by $p_T$. Therefore,
\[
\mathrm{KL}(p_T\|p_m)
=
\frac{1}{2}
e_m^\top F_T e_m
+
O(\|e_m\|^3).
\]

Teacher-relative KL can thus be interpreted locally as a Fisher-weighted squared logit error. This provides the connection to
the minimum-variance view above: $D_{\mathrm{Read}}$ and $D_{\mathrm{Reason}}$ act as empirical
proxies for the corresponding branch error magnitudes under this local
metric.

We define
\[
D_{\mathrm{Read}}=\mathbb{E}[\mathrm{KL}(p_T\|p_{\mathrm{Read}})],
\qquad
D_{\mathrm{Reason}}=\mathbb{E}[\mathrm{KL}(p_T\|p_{\mathrm{Reason}})],
\]
and, neglecting the cross-error term, obtain the reliability-based
estimate
\[
\hat{\lambda}
=
\frac{D_{\mathrm{Reason}}}{D_{\mathrm{Read}}+D_{\mathrm{Reason}}}.
\]

This substitution is an interpretation rather than an exact optimality
result. The local approximation need not hold when a branch is far from
the Teacher, and $D_{\mathrm{Read}}$ and $D_{\mathrm{Reason}}$ summarize marginal branch errors without accounting for their covariance.

\subsection{Teacher-Forced Fusion Diagnostics}

\begin{table}[t]
\centering
\small
\caption{
Reliability estimates and teacher-forced fusion diagnostics.
The teacher-KL optimum is the tested value of $\lambda$ that minimizes divergence from the full-context distribution.
}
\label{tab:fusion_diagnostics}
\begin{tabular}{lccccc}
\toprule
Benchmark &
$D_{\mathrm{Read}}$ &
$D_{\mathrm{Reason}}$ &
$\hat{\lambda}$ &
KL optimum &
Deployed $\lambda$ \\
\midrule
HotpotQA  & 0.385 & 0.882 & 0.70 & 0.75 & 0.70 \\
MuSiQue   & 0.476 & 0.963 & 0.67 & 0.70 & 0.70 \\
LooGLE    & 0.410 & 1.255 & 0.75 & 0.75 & 0.70 \\
LoCoMo & 0.273 & 1.325 & 0.83 & 0.85 & 0.85 \\
\bottomrule
\end{tabular}
\end{table}

The teacher-forced optimum follows the same broad pattern as the reliability estimate. LoCoMo favors the most Reader-heavy mixture, while the other benchmarks favor values near $0.70$--$0.75$. Exact agreement is not expected because teacher-relative token divergence and free-running sequence-level F1 optimize different objectives.

\section{Benchmark Construction}
\label{app:benchmarks}

For HotpotQA and MuSiQue, we construct a 40-document context for each question. We retain all documents provided with the original example, including the supporting documents provided distractors (10 documents for HotpotQA and 20 for MuSiQue). We use the raw question as a BM25 query to retrieve additional documents from the corresponding corpus until the context contains 40 documents. We then score all 40 documents with BM25 and order them by score before constructing the model input. This procedure adds and reorders distractor context without removing any document supplied with the original example.

For LooGLE and LoCoMo, we modify the standard independent-question evaluation into an accumulated setting by grouping questions that share the same source context into one episode. The source context is provided once, and the associated questions are asked sequentially while the context state persists across turns. This setting reflects interactive workloads such as chat and agents, where the same context needs to support multiple future questions rather than only the current one.

\subsection{HotpotQA}

We evaluate HotpotQA in a single-question setting. Each example contains 40 documents. The gold supporting documents are always included, and BM25 retrieval is used to select the remaining documents.

The full accuracy evaluation contains 600 examples. Examples are selected using a fixed random ordering with seed 42, which is held constant across all methods and operating points.

\subsection{MuSiQue}

We evaluate MuSiQue in the same single-question long-context setting. Each example contains 40 documents, with all gold documents included and the remaining documents selected by BM25 retrieval.

The evaluation contains 2,417 examples. The same examples and document ordering are used for every method.

\subsection{LooGLE}

Our LooGLE evaluation is derived from the \texttt{longdep\_qa} test split and uses an accumulated-context protocol. One document forms one conversation. The full document is supplied on the first turn, followed by the official questions associated with that document in their dataset order.

We remove examples contaminated by multiple-choice formatting and examples whose gold answer is a bare value of at most two digits. We additionally remove gold answers longer than 12 words, retain documents between 8,000 and 30,000 tokens, and limit each document to at most 30 questions.

After filtering, the evaluation contains 355 questions from 70 conversations. Conversations contain between 1 and 14 questions, with a median of 5. All retained questions are answerable from their associated document.

\subsection{LoCoMo}

For LoCoMo, we use an accumulated-context stress-test protocol derived from LoCoMo rather than the standard LoCoMo evaluation setting.

We use 10 conversations with 30 questions per conversation, yielding 300 evaluated questions. Questions within an episode share the same original context, which is provided only on the first turn. Subsequent turns reuse the accumulated state rather than re-supplying the original context.

Evaluation is performed at the question level. The order of questions within each conversation is fixed and is treated as part of the evaluation protocol.

\subsection{Evaluation Metric}

We report F1 on the answer string for all four benchmarks. Predictions and reference answers are lowercased, stripped of articles and punctuation, and compared as bags of tokens. We compute token-level precision and recall and report their harmonic mean. When multiple reference answers are available, we take the maximum F1 across references.

Our normalization follows the SQuAD-style answer evaluation with two modifications applied consistently to all methods. Hyphens are replaced with spaces rather than removed, so that expressions such as ``Tulip-tree'' and ``tulip tree'' match. Commas between digits are also replaced with spaces, so that answers such as ``2,4,1,3'' and ``2, 4, 1, 3'' are treated consistently.

This F1 corresponds to the standard answer metric for HotpotQA and MuSiQue and to the partial-match F1 used for LoCoMo QA. For LooGLE, we apply the same F1 metric to the long-dependency QA setting for consistency across benchmarks. LooGLE does not use this F1 as its original evaluation metric. Our accumulated LooGLE protocol and its evaluation metric therefore differ from the benchmark's standard setting.

\section{Generation and Accumulated Evaluation Details}
\label{app:generation_protocol}

\subsection{Generation and Answer Extraction}

All four primary benchmarks use the same prompting, decoding, answer-extraction, and scoring procedure. The maximum generation length is 200 tokens, with no separate token budget imposed on intermediate reasoning.

The prompt instructs the model to generate intermediate reasoning followed by the marker \texttt{Final Answer:}. Only the text extracted after this marker is used for evaluation. The reasoning tokens are part of the autoregressive trajectory but are not themselves scored.

After \texttt{Final Answer:} appears, decoding terminates once either the configured post-answer ramble boundary is detected or at least 24 words have been generated following the marker. Generation also terminates if a chat-leak marker is detected. If neither condition occurs, decoding continues until the 200-token maximum is reached.

If the generated output never contains \texttt{Final Answer:}, we use the first line of the raw output as the extracted answer.

PAIR, the full-context models, and all long-context baselines use the same generation and answer-extraction procedure.

\subsection{Single-Question Protocol}

HotpotQA and MuSiQue are evaluated as independent single-question episodes. Let $x$ denote the complete long context, $q$ the current question, and $y_{<t}$ the generated response prefix before decoding position $t$.

For PAIR, the Reader conditions on the full document, question, and generated prefix,

\[
p_{\mathrm{Read}}(y_t \mid x,q,y_{<t}),
\]

while the Reasoner does not receive the document and instead conditions on

\[
p_{\mathrm{Reason}}(y_t \mid q,y_{<t}).
\]

The fixed system instruction is omitted from the notation for clarity but is supplied to both branches.

At every decoding position, the Reader and the Reasoner receive the same previously generated response prefix. Their logits are fused as described in \Cref{sec:method}, and a single token is selected from the fused distribution. This token is then appended to the shared response prefix seen by both branches at the next decoding position. Thus, the branches do not generate separate reasoning trajectories.

For the Reader, the complete long context is prefetched into its KV state before response generation. The Reasoner never processes or stores document tokens. Each question begins a new episode, so no state or interaction history is carried between different HotpotQA or MuSiQue examples.

\subsection{Accumulated-Context Protocol}

LooGLE and LoCoMo are evaluated as multi-question episodes over a shared source context. Let $x$ denote the original long context and let $q_1,\ldots,q_K$ denote the sequence of questions associated with that context. Question order is fixed within each episode and is identical for all evaluated methods.

Let $H_i$ denote the committed interaction history after turn $i$, with

\[
H_0=\varnothing.
\]

At turn $i$, the Reader conditions on the original context, the committed history from preceding turns, the current question, and the response prefix generated so far:

\[
p_{\mathrm{Read}}(y_{i,t}\mid x,H_{i-1},q_i,y_{i,<t}).
\]

The Reasoner is reset at the beginning of every question and does not receive either the original document or previous interaction history:

\[
p_{\mathrm{Reason}}(y_{i,t}\mid q_i,y_{i,<t}).
\]

As in the single-question setting, both branches share the same generated prefix within the current response, and each next token is selected from their fused prediction.

After turn $i$ completes, we decompose its generated response into intermediate reasoning $r_i$ and its generated final answer. Before adding the turn to the persistent history, we replace the generated final answer with the reference answer $a_i^\star$:

\[
H_i
=
H_{i-1}
\oplus
(q_i,r_i,a_i^\star).
\]

Consequently, subsequent turns observe the model's previously generated reasoning but the correct previous final answer. This prevents an incorrect answer at one turn from changing the semantic facts presented to later turns, allowing the accumulated evaluation to measure persistent-context behavior without conflating it with cascading answer errors.

The original long context is supplied to the Reader only at the beginning of the episode. Its resulting KV state is then retained and extended as subsequent questions and committed histories are processed. The document is not re-prefilled from scratch before each question.

This distinction is important for query-dependent context-reduction methods. Let $S_i$ denote the subset of original document positions that remains represented in the persistent state after turn $i$, with

\[
S_0=\{1,\ldots,N\}.
\]

A reducer that updates persistent state based on the current question can be represented as

\[
S_i=\mathcal{R}(S_{i-1},q_i)\subseteq S_{i-1}.
\]

If a document position $j$ is removed at turn $i$, then without reprocessing the original context it cannot reappear in a later persistent state:

\[
j\notin S_i
\quad\Longrightarrow\quad
j\notin S_k
\qquad \text{for } k>i.
\]

A token that appears irrelevant to an early question can therefore be unavailable when a later question requires it. PAIR does not perform query-conditioned token removal from the Reader state and retains the complete document representation throughout the episode, corresponding to $S_i=S_0$ for all turns.

\subsection{Prompting and Output Format}
\label{app:prompts}

We instruct all models to produce a short reasoning step before the final answer so that intermediate evidence can appear in the generated prefix. HotpotQA, MuSiQue, and LooGLE use the following instruction.

\begin{Verbatim}[breaklines=true,fontsize=\small]
You are given context and a question. Answer in two steps and do BOTH.
Step 1 (Reasoning): 1-2 sentences using the context to work out the answer (who/what/when).
Step 2 (Answer): on a new line write 'Final Answer:' then ONLY the answer in a few words.

Rules:
- ALWAYS commit to one concrete best answer. NEVER reply 'Not mentioned', 'not provided', 'unavailable', or hedge — even if unsure, give your single most likely answer from the context.
- After 'Final Answer:' put ONLY the answer span: no notes, no parentheticals, no caveats, no extra sentence.
- No angle-bracket placeholders; don't repeat these instructions. Output nothing after the 'Final Answer:' line.
\end{Verbatim}

LoCoMo uses the same two-step format, with additional instructions for resolving dates across conversation sessions.

\begin{Verbatim}[breaklines=true,fontsize=\small]
You are answering a question about a long multi-session conversation (each session is dated). Answer in two steps.
Step 1 (Reasoning): ONE sentence — which turn/session answers it, and for a 'when' question resolve the actual date (e.g. if someone says they did it 'yesterday' in a session dated 8 May 2023, the answer is 7 May 2023).
Step 2 (Answer): on a new line write 'Final Answer:' then ONLY the short answer (a date, name, place, or brief phrase).

Rules:
- ALWAYS commit to your single best answer. NEVER reply 'Not mentioned' or 'not provided' — give the most likely answer.
- After 'Final Answer:' put ONLY the answer: no notes, parentheticals, or caveats.
- Phrase 'when' answers like the ground truth (e.g. '7 May 2023', 'the week before 9 June 2023').
- No angle-bracket placeholders; output nothing after the 'Final Answer:' line.
\end{Verbatim}

The same benchmark-specific instructions are used for Teacher generation, fine-tuning, and evaluation. Exact prompt assembly, accumulated-history formatting, and model-specific chat templates are provided in the released code.

\section{Baseline Configurations}
\label{app:baselines}

\subsection{Reference Models}

The full-context 32B Teacher and the 7B Reader define the two reference endpoints used throughout the paper. The Teacher processes the complete long context with the 32B model and retains its full long-context KV state. The Reader processes the same context with the 7B model and retains the corresponding 7B KV state.

All baseline methods use the same benchmark examples, prompts, generation limits, answer-extraction procedure, and F1 evaluation as Ours. For accumulated benchmarks, all methods are evaluated under the same persistent multi-turn protocol described in \Cref{app:generation_protocol}.

\subsection{Token-Reduction Baselines}

We evaluate SnapKV, ExpectedAttention, and SpecPrefill as representative methods that reduce the amount of long-context state processed or retained by the large model.

SnapKV and ExpectedAttention reduce the long-context KV state by selecting a subset of context tokens to retain. SpecPrefill instead reduces the context presented to the large model before full prefill. Although these methods intervene at different points in inference, all three ultimately trade context coverage for lower large-model memory or computation.

For each method, we evaluate the same nominal retained-context grid:
\[
\{15,\ 21.875,\ 30,\ 40\}\%.
\]
The 21.875\% point is included because it matches the ratio between the fp16 per-token KV footprints of the Qwen2.5 7B and 32B models:
\[
\frac{56\ \mathrm{KiB/token}}
{256\ \mathrm{KiB/token}}
=21.875\%.
\]
It therefore provides a useful reference operating point for comparing token reduction with the Reader-side representation used by Ours.

The nominal percentage is used only to configure the baseline. The memory coordinate shown in the paper is always obtained from the actual resident tensors and can therefore differ from the nominal retention rate.

\subsection{KV memory Accounting}

For all accuracy-memory comparisons, we measure resident KV cache bytes using the same tensor-level probe described in \Cref{app:memory}. The probe directly sums the storage occupied by the committed cache representation instead of inferring memory from the number of retained tokens.

Consequently, the x-coordinate of a baseline operating point includes any cache state that remains resident in addition to the nominally retained document tokens. For example, a configuration requested at 21.875\% token retention is not assumed to occupy exactly 21.875\% of the Teacher's KV memory.

The same accounting rule is used for our PAIR, the Teacher, the Reader, and the compression baselines. For accumulated benchmarks, memory is measured at question boundaries and averaged over the evaluation states used by the resident-memory probe.

\subsection{KV Quantization Baselines}

We evaluate KV quantization separately from token reduction because it changes the representation cost of the retained cache rather than selecting fewer context tokens.

For the full-context Teacher, we evaluate int8, int4, and int2 KV representations. Quantized KV-memory measurements include the packed KV representation, the fp16 residual where applicable, and all scale, zero-point, or other metadata required by the quantizer. The resulting memory fractions are therefore measured quantities rather than idealized bit-width ratios.

We additionally evaluate quantization composed with Ours by quantizing the Reader-side KV while leaving the Reasoner unchanged. These composition experiments test whether the architectural reduction of Ours and lower-bit KV storage provide complementary gains.

The Qwen2.5 quantization experiments use the HQQ-based cache implementation described in \Cref{app:quantization}. The detailed quantizer configuration, append-only accumulated-cache implementation, and matched-byte experiments are reported there.

\subsection{Common Evaluation Protocol}

All operating points within a benchmark use the same evaluation examples and generation protocol. The token-reduction budget, quantization precision, or model configuration is the only intended difference between operating points within the corresponding comparison.

The primary accuracy-memory figures use canonical F1 from the full evaluation set together with KV-memory measurements from the resident probe. Throughput is evaluated separately using the timing protocol in \Cref{app:throughput}; memory operating points should therefore not be interpreted as throughput measurements.

For accumulated benchmarks, context-reduction methods are evaluated under the same sequential question order as Ours. The original document is supplied according to the benchmark protocol, and the resulting KV cache is carried across turns. This is important because a token removed from KV cache for an earlier question may become relevant to a later question.

\section{Efficiency Evaluation Details}
\label{app:efficiency}

\subsection{KV Memory Measurement}
\label{app:memory}

The memory axis in \Cref{fig:main_efficiency} measures resident KV tensor storage rather than peak GPU memory. Its purpose is to quantify the model state that must remain resident for a sequence after context processing.

For each Transformer layer, we directly measure the storage occupied by tensors representing the committed key and value cache. For an fp16 cache, this consists of the key and value tensors. For quantized caches, we additionally include the packed quantized representation and all scale, zero-point, or other metadata required to reconstruct the cache. We compute the total from the resident tensors rather than estimating it from the nominal number of retained tokens.

For PAIR, the measurement includes both branches. The first component is the long-context KV stored by the 7B Reader. The second is the Reasoner KV associated with its instruction, current query, and generated prefix. Consequently, the measured footprint of PAIR is slightly larger than the per-token KV ratio between the 7B and 32B models.

For accumulated benchmarks, KV memory is measured after each evaluated question and averaged over questions. The memory probe processes one conversation at a time. Batch padding is excluded because the metric is intended to characterize persistent state per sequence rather than temporary storage introduced by a particular serving batch.

For Qwen2.5, the fp16 KV footprint is approximately 256 KiB per token for the 32B model and 56 KiB per token for the 7B model. The corresponding per-token ratio is 21.875\%. This value provides a useful reference point for the memory footprint of PAIR, although its measured value is slightly higher because of the Reasoner's short-context KV.

\subsection{Operating Points Reported in the Main Figures}

\Cref{tab:baseline_operating_points} summarizes the configurations used in the main accuracy--memory comparison. All non-quantized configurations use fp16 KV. Nominal budgets are reported for token-reduction methods, while all plotted memory coordinates use measured resident KV.

\begin{table}[t]
\centering
\small
\caption{Configurations evaluated in the accuracy--memory comparison.}
\label{tab:baseline_operating_points}
\begin{tabular}{ll}
\toprule
Method & Evaluated settings \\
\midrule
Full-context 32B & full context \\
Reader 7B & full context \\
PAIR (32B + 7B) & full Reader context \\
SnapKV & 15, 21.875, 30, 40\% \\
ExpectedAttention & 15, 21.875, 30, 40\% \\
SpecPrefill & 15, 21.875, 30, 40\% \\
32B KV quantization & int8, int4, int2 \\
\bottomrule
\end{tabular}
\end{table}

\subsection{Raw Accuracy-Memory Results}

The exact F1 and measured KV-memory footprint of every operating point shown in the main figures are reported in \Cref{tab:raw_accuracy_memory}. We report measured memory rather than nominal token-retention rate so that the plotted points can be reproduced directly from the tabulated values.

For token-reduction methods, the table should contain both the nominal configuration budget and the measured resident-KV percentage. For quantized configurations, the table should report the measured resident footprint including quantization metadata.

\begin{table*}[t]
\centering
\scriptsize
\caption{
Raw accuracy--memory operating points across the four primary benchmarks.
All non-quantized configurations use fp16 KV.
}
\label{tab:raw_accuracy_memory}

\setlength{\tabcolsep}{4.5pt}
\begin{tabular}{llcccccccc}
\toprule
& &
\multicolumn{2}{c}{HotpotQA} &
\multicolumn{2}{c}{MuSiQue} &
\multicolumn{2}{c}{LooGLE} &
\multicolumn{2}{c}{LoCoMo} \\
\cmidrule(lr){3-4}
\cmidrule(lr){5-6}
\cmidrule(lr){7-8}
\cmidrule(lr){9-10}
Method & Setting &
KV / 32B & F1 &
KV / 32B & F1 &
KV / 32B & F1 &
KV / 32B & F1 \\
\midrule

Full-context 32B & full
& 100.0\% & 0.7492
& 100.0\% & 0.4695
& 100.0\% & 0.4876
& 100.0\% & 0.5583 \\

Reader 7B & full
& 21.9\% & 0.5939
& 21.9\% & 0.3323
& 21.9\% & 0.3878
& 21.9\% & 0.4243 \\

PAIR (32B + 7B) & --
& 24.8\% & 0.7178
& 25.2\% & 0.4380
& 22.7\% & 0.4438
& 23.2\% & 0.4992 \\

\midrule

SnapKV & 5\%
& 6.7\% & 0.6163
& 6.9\% & 0.2913
& 6.6\% & 0.2836
& 13.0\% & 0.2756 \\

SnapKV & 10\%
& 11.6\% & 0.6564
& 11.8\% & 0.3436
& 11.6\% & 0.3226
& 18.0\% & 0.3157 \\

SnapKV & 15\%
& 16.5\% & 0.6708
& 16.7\% & 0.3669
& 16.5\% & 0.3280
& 22.0\% & 0.3703 \\

SnapKV & 21.875\%
& 23.3\% & 0.7027
& 23.5\% & 0.3874
& 23.2\% & 0.3606
& 28.7\% & 0.3875 \\

SnapKV & 30\%
& 31.3\% & 0.7016
& 31.4\% & 0.4004
& 31.2\% & 0.3668
& 36.1\% & 0.4320 \\

SnapKV & 40\%
& 41.2\% & 0.7084
& 41.2\% & 0.4166
& 41.1\% & 0.4123
& 45.3\% & 0.4937 \\

\midrule

ExpectedAttention & 5\%
& 7.7\% & 0.4410
& 8.1\% & 0.1815
& 6.9\% & 0.2559
& 13.3\% & 0.2702 \\

ExpectedAttention & 10\%
& 12.6\% & 0.5171
& 13.0\% & 0.2131
& 11.8\% & 0.2606
& 17.8\% & 0.2819 \\

ExpectedAttention & 15\%
& 17.5\% & 0.5701
& 17.8\% & 0.2406
& 16.7\% & 0.2805
& 22.8\% & 0.2964 \\

ExpectedAttention & 21.875\%
& 24.1\% & 0.6061
& 24.5\% & 0.2798
& 23.4\% & 0.3129
& 29.0\% & 0.3269 \\

ExpectedAttention & 30\%
& 32.0\% & 0.6369
& 32.3\% & 0.3199
& 31.4\% & 0.3462
& 36.4\% & 0.3577 \\

ExpectedAttention & 40\%
& 41.7\% & 0.6646
& 42.0\% & 0.3449
& 41.2\% & 0.3628
& 45.5\% & 0.4136 \\

\midrule

SpecPrefill & 5\%
& 7.8\% & 0.3518
& 8.1\% & 0.1338
& 6.9\% & 0.3015
& 13.6\% & 0.2884 \\

SpecPrefill & 10\%
& 12.6\% & 0.5245
& 12.9\% & 0.2264
& 11.8\% & 0.3400
& 18.1\% & 0.2837 \\

SpecPrefill & 15\%
& 17.5\% & 0.6102
& 17.8\% & 0.2887
& 16.7\% & 0.3552
& 22.7\% & 0.3354 \\

SpecPrefill & 21.875\%
& 24.2\% & 0.6539
& 24.5\% & 0.3599
& 23.4\% & 0.3760
& 29.0\% & 0.3607 \\

SpecPrefill & 30\%
& 32.1\% & 0.7010
& 32.4\% & 0.3921
& 31.4\% & 0.3857
& 36.4\% & 0.3526 \\

SpecPrefill & 40\%
& 41.8\% & 0.7064
& 42.1\% & 0.4257
& 41.2\% & 0.4160
& 45.5\% & 0.4078 \\

\midrule

32B KV quant. & int8
& 53.9\% & 0.7625
& 53.9\% & 0.4793
& 53.5\% & 0.4927
& 53.6\% & 0.5631 \\

32B KV quant. & int4
& 28.6\% & 0.7386
& 28.5\% & 0.4571
& 28.5\% & 0.4828
& 28.6\% & 0.5584 \\

32B KV quant. & int2
& 16.8\% & 0.3994
& 17.0\% & 0.1761
& 16.0\% & 0.1538
& 16.0\% & 0.1729 \\

\bottomrule
\end{tabular}
\end{table*}

\subsection{Query-Dependent Compression Across Accumulated Turns}
\label{app:accumulated_compression}

The accumulated protocol exposes a limitation that does not appear in single-question evaluation. In our implementation, SnapKV and SpecPrefill compress the document cache once using the first question of an episode. The resulting compressed cache is then kept fixed for the remaining questions, and the original document is not re-prefilled. The first question is therefore the only turn for which the retained context is selected using the question being answered.

We measure this effect on LooGLE using the 65 conversations that contain at least two questions. To avoid differences in difficulty across conversations, we compare the first question with later questions within each conversation and average the paired differences. SnapKV drops from 0.455 F1 on the first question to 0.341 on later questions, for a paired difference of $-0.115$. SpecPrefill drops from 0.422 to 0.368, for a paired difference of $-0.051$. PAIR does not show this decline, increasing from 0.420 to 0.453. Full-context 32B similarly increases from 0.471 to 0.495.

\begin{table}[t]
\centering
\small
\caption{
First-question and later-question accuracy on the 65 LooGLE conversations with at least two questions. Differences are computed within conversation before averaging.
}
\label{tab:loogle_first_later}
\begin{tabular}{lccc}
\toprule
Method & First & Later & Later $-$ First \\
\midrule
Full-context 32B        & 0.471 & 0.495 & +0.042 \\
Full-context 7B         & 0.372 & 0.392 & +0.007 \\
PAIR                     & 0.420 & 0.453 & +0.043 \\
SnapKV                   & 0.455 & 0.341 & -0.115 \\
SpecPrefill              & 0.422 & 0.368 & -0.051 \\
ExpectedAttention        & 0.295 & 0.321 & +0.013 \\
KV quantization          & 0.438 & 0.497 & +0.059 \\
\bottomrule
\end{tabular}
\end{table}

The first-to-later drop is specific to the methods whose persistent state is selected using the first query. The other methods do not show the same pattern. This comparison is paired within conversation because LooGLE conversation length is correlated with difficulty. For example, the Teacher's first-question F1 is 0.319 for conversations with at least five questions and 0.468 for shorter conversations, so an unpaired comparison across question positions would be confounded by conversation composition.

The same issue can occur when evidence needed by a later question was not selected for the first one. In one LooGLE episode, the first question asks which two regions of Asia are associated with the stories, while a later question asks how many stories are reported by Jan-Öjvind Swahn. The full-context model answers the later question correctly, while SnapKV receives zero F1 on the gold answer ``3 stories.'' This example illustrates the general failure mode rather than serving as a separate quantitative result.

LoCoMo shows the same qualitative direction, but its position-wise results require more care because question categories are not uniformly distributed across turns and the evaluation contains only ten conversations. We therefore control for question category when comparing early and later turns and use LoCoMo only as supporting evidence rather than as a second independent estimate of the LooGLE effect.

Single-question benchmarks do not expose this persistence mismatch. On HotpotQA and MuSiQue, the compressed state only needs to answer the same query that was used to construct it. Accumulated evaluation instead tests whether a persistent context representation remains useful for questions that were not known when that state was created.

\subsection{Timing Workloads and Batch Selection}
\label{app:throughput}
Throughput is measured on fixed timing workloads rather than on the full accuracy evaluation sets. We use 8 conversations comprising 240 answers for LoCoMo, 48 conversations comprising 261 answers for LooGLE, 96 questions for MuSiQue, and 96 questions for HotpotQA. Every configuration within a benchmark receives the identical timing workload.

For each configuration, we evaluate a batch-size ladder~\citep{orthrus} and report the largest tested batch that completes without running out of memory. We refer to this as the largest tested feasible batch rather than the exact maximum batch because intermediate batch sizes are not exhaustively searched.

For accumulated benchmarks, batching is performed over conversations and the original question order within each conversation is preserved. When the workload ends with a partial batch, we record both throughput over the complete workload and throughput over complete batches only. For the main accuracy--cost frontier, we use the lower rate for PAIR and the higher rate for competing configurations, making the comparison conservative with respect to PAIR. The exact rate used for every plotted point is reported below.

All configurations contributing to a benchmark-level comparison are measured on the same GPU node. Measurements obtained on a different node are not inserted into an existing throughput frontier.

\subsection{Execution Paths}
\label{app:fkv}

The main all-method accuracy--throughput comparison uses the standard Hugging Face autoregressive decoding path used by each implementation. Full-context 32B, Full-context 7B, and PAIR use \texttt{DynamicCache}, while SnapKV, ExpectedAttention, and SpecPrefill retain their method-specific cache transformations. We use this path for the all-method comparison because the token-reduction baselines modify or compact the KV cache and are not implemented on our fixed-KV backend; accelerating only the dense configurations would confound the architectural comparison with an execution-backend advantage.

The standard \texttt{DynamicCache} path, however, incurs substantial cache-management overhead during decoding because the KV state grows dynamically, requiring repeated cache growth and copying. This overhead becomes increasingly significant at long contexts and larger batches and can obscure the underlying scaling behavior of the model architectures. In a representative profile, KV-cache growth and copying accounted for 11.47 ms of a 30.27 ms decode step.

For the controlled batch- and context-scaling experiments, we therefore use the same fixed-KV (FKV) backend for Full-context 32B, Full-context 7B, and PAIR. FKV preallocates per-row KV buffers, updates them in place, uses \texttt{flash\_attn\_with\_kvcache} for decoding, and captures the one-token decode step with CUDA Graphs. By removing repeated cache growth and copying, FKV allows these experiments to more directly expose the computational and memory scaling of the three architectures.

\subsection{Raw Accuracy--Throughput Operating Points}

\Cref{tab:raw_throughput} reports the raw throughput measurements for every operating point shown in the bottom row of \Cref{fig:main_efficiency}. ``All'' denotes throughput over the complete timing workload, while ``Complete'' excludes a final partial batch when present. HotpotQA and MuSiQue use 96-question workloads on nodes gh105 and gh126, respectively; LooGLE uses a 48-conversation, 261-answer workload on gh042; and LoCoMo uses an 8-conversation, 240-answer workload on gh065.

\begin{table*}[t]
\centering
\scriptsize
\caption{
Raw accuracy--throughput operating points across the four primary benchmarks.
Throughput is measured in completed answers per second.
}
\label{tab:raw_throughput}

\setlength{\tabcolsep}{2.7pt}
\begin{tabular}{llcccccccccccc}
\toprule
& &
\multicolumn{3}{c}{HotpotQA} &
\multicolumn{3}{c}{MuSiQue} &
\multicolumn{3}{c}{LooGLE} &
\multicolumn{3}{c}{LoCoMo} \\
\cmidrule(lr){3-5}
\cmidrule(lr){6-8}
\cmidrule(lr){9-11}
\cmidrule(lr){12-14}
Method & Setting &
Batch & All & Complete &
Batch & All & Complete &
Batch & All & Complete &
Batch & All & Complete \\
\midrule

Full-context 32B & full
& 12 & 0.560 & 0.560
& 10 & 0.542 & 0.547
& 3 & 0.154 & 0.154
& $\geq4$ & 0.347 & 0.347 \\

Full-context 7B & full
& 80 & 1.712 & 1.760
& 48 & 2.302 & 2.302
& 16 & 0.458 & 0.458
& $\geq8$ & 1.159 & 1.159 \\

PAIR (32B + 7B) & --
& 20 & 1.003 & 0.993
& 16 & 1.041 & 1.041
& 6 & 0.238 & 0.238
& $\geq8$ & 0.614 & 0.614 \\

\midrule

SnapKV & 5\%
& $\geq96$ & 0.880 & 0.880
& $\geq96$ & 0.963 & 0.963
& 24 & 0.443 & 0.443
& $\geq8$ & 0.874 & 0.874 \\

SnapKV & 10\%
& $\geq80$ & 0.749 & 0.810
& $\geq80$ & 0.856 & 0.863
& 16 & 0.375 & 0.375
& $\geq8$ & 0.884 & 0.884 \\

SnapKV & 15\%
& 48 & 0.797 & 0.797
& 48 & 0.886 & 0.886
& 10 & 0.295 & 0.312
& $\geq8$ & 0.803 & 0.803 \\

SnapKV & 21.875\%
& 40 & 0.646 & 0.679
& 32 & 0.806 & 0.806
& 8 & 0.256 & 0.256
& $\geq8$ & 0.817 & 0.817 \\

SnapKV & 30\%
& 24 & 0.647 & 0.647
& 24 & 0.767 & 0.767
& 6 & 0.221 & 0.221
& $\geq8$ & 0.723 & 0.723 \\

SnapKV & 40\%
& 20 & 0.591 & 0.605
& 20 & 0.733 & 0.720
& 4 & 0.197 & 0.197
& 6 & 0.415 & 0.537 \\

\midrule

ExpectedAttention & 5\%
& $\geq96$ & 0.826 & 0.826
& $\geq96$ & 0.915 & 0.915
& 24 & 0.357 & 0.357
& $\geq8$ & 0.854 & 0.854 \\

ExpectedAttention & 10\%
& $\geq80$ & 0.733 & 0.796
& $\geq80$ & 0.805 & 0.846
& 16 & 0.339 & 0.339
& $\geq8$ & 0.845 & 0.845 \\

ExpectedAttention & 15\%
& 48 & 0.792 & 0.792
& 48 & 0.819 & 0.819
& 10 & 0.262 & 0.270
& $\geq8$ & 0.847 & 0.847 \\

ExpectedAttention & 21.875\%
& 40 & 0.668 & 0.670
& 32 & 0.781 & 0.781
& 8 & 0.220 & 0.220
& $\geq8$ & 0.857 & 0.857 \\

ExpectedAttention & 30\%
& 24 & 0.641 & 0.641
& 24 & 0.738 & 0.738
& 6 & 0.194 & 0.194
& $\geq8$ & 0.778 & 0.778 \\

ExpectedAttention & 40\%
& 20 & 0.609 & 0.589
& 20 & 0.724 & 0.716
& 4 & 0.166 & 0.166
& 6 & 0.485 & 0.600 \\

\midrule

SpecPrefill & 5\%
& $\geq48$ & 0.798 & 0.798
& $\geq48$ & 0.756 & 0.756
& 16 & 0.483 & 0.483
& $\geq8$ & 1.177 & 1.177 \\

SpecPrefill & 10\%
& $\geq40$ & 0.659 & 0.704
& $\geq40$ & 0.784 & 0.813
& 10 & 0.391 & 0.412
& $\geq8$ & 0.999 & 0.999 \\

SpecPrefill & 15\%
& 32 & 0.707 & 0.707
& 24 & 0.714 & 0.714
& 6 & 0.292 & 0.292
& $\geq8$ & 0.885 & 0.885 \\

SpecPrefill & 21.875\%
& $\geq24$ & 0.667 & 0.667
& 20 & 0.650 & 0.646
& 4 & 0.236 & 0.236
& 6 & 0.541 & 0.711 \\

SpecPrefill & 30\%
& 16 & 0.549 & 0.549
& 12 & 0.508 & 0.508
& 3 & 0.218 & 0.218
& 4 & 0.471 & 0.471 \\

SpecPrefill & 40\%
& 12 & 0.456 & 0.456
& 8 & 0.405 & 0.405
& 2 & 0.172 & 0.172
& 4 & 0.458 & 0.458 \\

\bottomrule
\end{tabular}
\end{table*}

\section{Two-GPU Branch-Parallel Execution}
\label{app:two_gpu}
PAIR does not require multiple GPUs, but the Reader and Reasoner interact only when their logits are fused and can therefore execute in parallel when separate devices are available. We evaluate this deployment setting on LooGLE by placing the 7B Reader and 32B Reasoner on separate GPUs.

Parallel execution improves both latency and aggregate throughput. Relative to one-GPU PAIR, the two-GPU configuration increases throughput by 2.07$\times$, from 0.59 to 1.22 answers/s. This is also slightly higher than the 1.18 answers/s obtained by using the same two GPUs for two independent PAIR replicas, showing that reducing request latency through Reader--Reasoner parallelism does not sacrifice aggregate serving capacity. Compared with two full-context large model replicas, parallel PAIR provides 1.58$\times$ higher throughput.

The latency benefit is largest when long-context processing dominates. The first LooGLE answer, which includes document prefill, takes 8.31\,s with parallel PAIR compared with 12.10\,s for the full-context large model, a 1.46$\times$ speedup. On later turns, where the persistent context is already available and decoding dominates, latency is similar at 2.34\,s versus 2.47\,s. Thus, parallel Reader--Reasoner execution provides a practical path to lower request latency while retaining the throughput advantage of replication.

\subsection{Execution Setup}

We evaluate two-GPU execution on LooGLE-accum. The 7B Reader and its long-context KV state are placed on GPU 0, while the 32B Reasoner is placed on GPU 1. The Reader and the Reasoner execute concurrently during generation and synchronize only when their logits are fused.

The experiment uses two NVIDIA GH200 GPUs on the same node. The dense decode path uses FKV with CUDA Graphs. Long-document prefill is performed entirely by the Reader because the Reasoner never receives the document. Reader prefill is divided into 1,024-token chunks to control activation memory.

The one-GPU and two-GPU configurations produce byte-identical answers on the matched validation run. The comparison therefore changes only execution scheduling and hardware placement, not model outputs.

\subsection{Aggregate Throughput}

The throughput comparison uses the same 40 LooGLE conversations containing 213 answers. The one-GPU Ours operating point is batch 4, while the two-GPU branch-parallel configuration reaches batch 16.

\begin{table}[t]
\centering
\small
\caption{Aggregate throughput for one- and two-GPU execution on LooGLE.}
\label{tab:two_gpu_throughput}
\begin{tabular}{lccc}
\toprule
Configuration & GPUs & Batch & Answers/s \\
\midrule
Ours, one GPU & 1 & 4 & 0.5896 \\
Two independent one-GPU Ours replicas & 2 & 4 each & 1.1793 \\
Ours, branch-parallel & 2 & 16 & \textbf{1.2195} \\
Full-context large model, one GPU & 1 & 2 & 0.3868 \\
Two independent full-context large model replicas & 2 & 2 each & 0.7736 \\
\bottomrule
\end{tabular}
\end{table}

Relative to one-GPU Ours, branch-parallel execution increases aggregate throughput by 2.068$\times$. Relative to two independent one-GPU Ours replicas, the throughput per GPU is 1.034$\times$. At equal two-GPU resources, branch-parallel Ours provides 1.576$\times$ the aggregate throughput of two full-context large model replicas.

\subsection{Single-Request Latency}

Throughput-optimal batching is not appropriate for evaluating request latency. We therefore report latency separately at batch 4 for two-GPU PAIR and compare it with the lower-latency one-GPU operating points.

\begin{table}[t]
\centering
\small
\caption{Per-request latency on LooGLE. The first answer includes long-document prefill; later turns reuse the accumulated context state.}
\label{tab:two_gpu_latency}
\begin{tabular}{lccc}
\toprule
Configuration & Batch & First answer & Later turn \\
\midrule
Full-context large model, one GPU & 2 & 12.10\,s & 2.47\,s \\
PAIR, one GPU & 4 & 9.24\,s & 3.14\,s \\
PAIR, two GPUs & 4 & \textbf{8.31\,s} & \textbf{2.34\,s} \\
\bottomrule
\end{tabular}
\end{table}

The first answer includes long-document prefill, which gives PAIR a large advantage because the document is processed by the 7B Reader rather than the full-context 32B. Two-GPU PAIR is therefore 1.457$\times$ faster than the full-context large model on the first answer.

Later turns reuse the persistent context state and require little additional prefill, so latency is dominated by autoregressive decoding. In this regime, the Reader and the Reasoner perform different amounts of work at each decoding step and must synchronize for logit fusion, which limits the benefit of branch parallelism. Despite this less favorable regime, two-GPU PAIR remains 1.056$\times$ faster than the full-context 32B on later turns.

The latency experiment uses batch size as 4, while the aggregate-throughput result uses batch 16. The smaller batch is used for latency because increasing batch size improves aggregate throughput but increases the wall-clock latency of an individual request.

\subsection{Full-context large model Replication and Sharding}
For aggregate throughput, the natural use of a second GPU for the full-context 32B model is independent replication. We also tested sharding a single full-context 32B across both GPUs. In this setting, sharding increased the feasible batch size but substantially reduced throughput: the sharded full-context 32B model reached 0.1121 answers/s, compared with 0.3326 answers/s for two independent replicas under the corresponding comparison.

The sharded and replicated full-context 32B model measurements use the standard Hugging Face execution path because the CUDA Graph used by FKV does not span multiple devices. These measurements are used only to establish that independent replication is the stronger full-context 32B model throughput strategy in our environment.

\section{Cross-Family Evaluation Details}
\label{app:model_generalization}

The cross-family experiments use the same HotpotQA context construction as the primary Qwen2.5 experiments. Every model pair is evaluated on the same 600 questions at 40, 80, 120, and 160 documents, corresponding to approximately 6.1k, 11.8k, 18.0k, and 23.3k measured context tokens. Full-context large model, Ours, and the small-model baseline within each model pair are always evaluated on identical examples.

Each model family uses its own full-context large model(Teacher) to generate training targets and its own trained Reader and Reasoner adapters. The fusion rule and semantic content of the task prompt are unchanged across model families. Prompt serialization follows the native chat format of each model family rather than forcing an identical token-level template.

For Gemma-3, all reported results use the model's native chat format consistently during Teacher-target generation, Reader training, Reasoner training, and evaluation. The semantic prompt, fusion weight, generation length, greedy decoding procedure, answer extractor, evaluation boundary, and HotpotQA examples are otherwise unchanged from the canonical protocol. Earlier Gemma experiments using a plain prompt are not included in the reported comparison.

\begin{table*}[t]
\centering
\small
\caption{
Raw HotpotQA F1 for the cross-family generalization experiment.
All configurations use the same 600 questions and context construction.
}
\label{tab:model_generalization_raw}
\begin{tabular}{llccc}
\toprule
Model pair & Context & Full-context Large model & PAIR (ours) & Full-context Reader model \\
\midrule
Qwen2.5 32B+7B
& 6.1k  & 0.7492 & 0.7178 & 0.5939 \\
& 11.8k & 0.7232 & 0.6885 & 0.5969 \\
& 18.0k & 0.7029 & 0.6828 & 0.5916 \\
& 23.3k & 0.6893 & 0.6718 & 0.5554 \\
\midrule
Qwen2.5 32B+3B
& 6.1k  & 0.7492 & 0.6708 & 0.5432 \\
& 11.8k & 0.7232 & 0.6550 & 0.5057 \\
& 18.0k & 0.7029 & 0.6309 & 0.4488 \\
& 23.3k & 0.6893 & 0.6243 & 0.4515 \\
\midrule
OLMo-3 32B+7B
& 6.1k  & 0.6465 & 0.5942 & 0.4710 \\
& 11.8k & 0.6110 & 0.5527 & 0.4402 \\
& 18.0k & 0.5930 & 0.5545 & 0.4086 \\
& 23.3k & 0.5855 & 0.5128 & 0.3776 \\
\midrule
Gemma-3 12B+4B
& 6.1k  & 0.6940 & 0.6588 & 0.4636 \\
& 11.8k & 0.6683 & 0.6449 & 0.4082 \\
& 18.0k & 0.6356 & 0.6244 & 0.4520 \\
& 23.3k & 0.6449 & 0.6330 & 0.4287 \\
\bottomrule
\end{tabular}
\end{table*}

\section{Scalability Evaluation Details}
\label{app:scalability}

\subsection{Batch-Scaling Protocol}
\label{app:batch_scaling}

The batch-scaling experiments use the same 96 HotpotQA questions at every context length. We evaluate four document depths corresponding to approximately 6.1k, 11.8k, 18.0k, and 23.3k measured context tokens.

At each context length, the full-context 32B model(reference), Reader, and PAIR are evaluated over increasing batch-size rungs until the next tested batch runs out of memory or the predefined sweep limit is reached. Every reported point measures completed answers per second over the full end-to-end workload using the FKV path.

The largest tested surviving batches for the full-context 32B model are 10, 6, 4, and 3 at 6.1k, 11.8k, 18.0k, and 23.3k tokens, respectively. For Ours, the corresponding batches are 16, 10, 6, and 6. For the 7B Reader, they are 64, 32, 24, and 24.

\subsection{Context-Length Scaling Details}
\label{app:context_scaling}

The context-length experiment uses the same HotpotQA examples at every context size. Accuracy is evaluated on the same 600-question set, and throughput is measured on the same 96-question timing workload. Context length is increased by adding retrieved documents while preserving the same evaluation questions.

The 40-, 80-, 120-, and 160-document settings correspond to measured context lengths of 6.1k, 11.8k, 18.0k, and 23.3k tokens and use the native Qwen2.5 positional configuration. Longer contexts exceed the native 32k context window, so from 200 documents onward we enable YaRN with scaling factor 4 for the full-context 32B model(reference), the PAIR, and the 7B Reader. No other model or decoding setting changes at this boundary.

Throughput is measured using the FKV execution path, with each configuration evaluated at its largest tested feasible batch size. \Cref{tab:context_throughput_raw} reports the exact operating points shown in \Cref{fig:context_scaling}.

\begin{table*}[t]
\centering
\scriptsize
\caption{Raw operating points for the context-length throughput experiment. Each method is evaluated at its largest tested feasible batch size using FKV.}
\label{tab:context_throughput_raw}
\begin{tabular}{lccccccc}
\toprule
Context & Positional setting &
Reference(32B) $B$ & Reference ans/s &
PAIR $B$ & PAIR ans/s &
Reader(7B) $B$ & Reader ans/s \\
\midrule
6.1k  & native          & 10 & 0.815 & 16 & 1.973 & 64 & 3.141 \\
11.8k & native          &  6 & 0.432 & 10 & 1.116 & 32 & 1.729 \\
18.0k & native          &  4 & 0.279 &  6 & 0.745 & 24 & 1.129 \\
23.3k & native          &  3 & 0.197 &  6 & 0.592 & 24 & 0.814 \\
29.2k & YaRN $\times 4$ &  2 & 0.145 &  4 & 0.415 & 16 & 0.621 \\
34.8k & YaRN $\times 4$ &  2 & 0.119 &  4 & 0.367 & 16 & 0.491 \\
40.9k & YaRN $\times 4$ &  2 & 0.100 &  4 & 0.311 & 12 & 0.411 \\
45.7k & YaRN $\times 4$ &  1 & 0.078 &  3 & 0.250 & 12 & 0.332 \\
\bottomrule
\end{tabular}
\end{table*}

\subsection{Accuracy Associated with Timing Measurements}

Task accuracy is always taken from the canonical full-set evaluation rather than from the smaller timing workload. The timing workloads are used only to measure execution cost and serving capacity.

FKV runs likewise do not define the task-accuracy values used in the accuracy-cost figures. Changes in attention kernels and KV-buffer allocation can introduce small floating-point differences that occasionally alter a greedy decoding decision. We therefore associate each FKV timing point with the corresponding canonical accuracy measured using the standard evaluation path.

\section{Combining PAIR with KV Quantization}
\label{app:quantization}

KV quantization reduces KV memory along a different axis from PAIR. Quantization preserves the same KV state but stores each element at lower precision, whereas PAIR reduces the size of the model that stores the long-context state. We therefore ask three questions: whether standalone quantization can match PAIR's accuracy--memory operating point, whether quantization can be applied on top of PAIR, and whether quantized token-reduction baselines change the comparison at similar cache-byte budgets.

Overall, the two approaches are complementary. Standalone 32B quantization provides only coarse memory operating points around PAIR: int4 preserves accuracy but remains above PAIR's KV memory footprint, while int2 reduces memory further but causes a large accuracy loss. In contrast, quantizing PAIR's Reader to int8 preserves its accuracy and further reduces KV memory. Quantized token-reduction baselines also remain less accurate than PAIR at similar cache-byte budgets.
\subsection{Quantization Setup and Memory Accounting}

The Qwen2.5 experiments use an HQQ-based KV cache because the official KIVI implementation does not directly support the Qwen2.5 configuration used in our experiments. Unless otherwise specified, we use group size 64, a 128-token fp16 residual, and quantization axis 0 for both keys and values.

For accumulated-context benchmarks, the quantized cache is append-only. Once a complete 128-token block is committed, that block is quantized once and remains immutable; only the newest residual remains in fp16. Rollback restores the previous committed-block list and residual rather than reconstructing and re-quantizing historical blocks. This avoids repeatedly quantizing historical state across turns. In a synthetic 30-turn probe, reconstruction error for a fixed committed block increased by 3.45$\times$ under repeated reconstruction and re-quantization, compared with 1.00$\times$ under the append-only implementation.

The current HQQ path reconstructs quantized KV into fp16 for attention and is not optimized for serving throughput. We therefore use quantized runs only for accuracy and KV-memory analysis and make no throughput claim from these experiments.

For the KV-memory measurements, we use the same resident-memory probe as in \Cref{sec:main_efficiency}. At batch size one, the probe directly sums the resident K/V tensors and all quantization metadata at answer boundaries. Quantized representations therefore include packed values, scales, zero points, and any residual fp16 state required by the cache. The same fixed probe conversations are used across methods within each benchmark: 24 examples for HotpotQA, 24 for MuSiQue, 60 question states for LoCoMo, and 17 conversations for LooGLE. The fp16 full-context large model denominator is measured on the same probe workload.

These resident measurements define the KV-memory percentages reported below. They are distinct from the cache-byte measurements used in the matched-byte experiments in \Cref{app:quantized_matched_byte}, which are obtained from batched accuracy runs, include padding to the longest row in each batch, and answer a different comparison.

\subsection{Standalone Full-Context Quantization}

We first ask whether quantizing the full-context 32B model provides an accuracy--memory operating point comparable to PAIR. Because the available precision levels are discrete, a direct matched-memory comparison is not possible.

\Cref{tab:quantization_main} shows that int4 preserves the 32B model's accuracy well but uses approximately 28.5\% of its fp16 KV memory, still above PAIR's 22.7--25.2\%. Int2 reduces memory below PAIR, to approximately 16--17\%, but causes a large accuracy drop on every benchmark. PAIR therefore lies between the two useful memory levels exposed by standalone 32B quantization: it uses less KV memory than the accuracy-preserving int4 configuration while avoiding the degradation of int2.

\begin{table}[!t]
\centering
\footnotesize
\caption{
F1 and resident KV memory for KV quantization of the full-context 32B model.
Parentheses show KV memory relative to the fp16 full-context model.
PAIR is included as a reference operating point.
}
\label{tab:quantization_main}
\begin{tabular}{lccccc}
\toprule
Benchmark &
32B fp16 &
32B int8 &
32B int4 &
32B int2 &
PAIR fp16 \\
\midrule
HotpotQA &
0.7492 (100\%) &
0.7625 (53.9\%) &
0.7386 (28.6\%) &
0.3994 (16.8\%) &
0.7178 (24.8\%) \\

MuSiQue &
0.4695 (100\%) &
0.4793 (53.9\%) &
0.4571 (28.5\%) &
0.1761 (17.0\%) &
0.4380 (25.2\%) \\

LooGLE &
0.4876 (100\%) &
0.4927 (53.5\%) &
0.4828 (28.5\%) &
0.1538 (16.0\%) &
0.4438 (22.7\%) \\

LoCoMo &
0.5583 (100\%) &
0.5631 (53.6\%) &
0.5584 (28.6\%) &
0.1729 (16.0\%) &
0.4992 (23.2\%) \\
\bottomrule
\end{tabular}
\end{table}

\subsection{Quantizing the Reader in PAIR}

The previous comparison treats quantization as an alternative to PAIR, but the two methods modify different aspects of the persistent state. We next ask whether quantization can instead be applied directly to PAIR. We quantize only the Reader-side KV while leaving the Reasoner unchanged.

As shown in \Cref{tab:reader_quantization}, Reader-int8 changes F1 by less than 0.01 on every benchmark. On LooGLE, where we additionally run the canonical resident-memory probe, it reduces PAIR's KV memory from 22.7\% to 14.3\% of the fp16 32B footprint while retaining 0.4364 F1. Thus, the memory reduction from KV quantization can be added on top of PAIR rather than treated solely as a competing alternative.

\begin{table}[t]
\centering
\small
\caption{
Effect of applying int8 KV quantization to the Reader in PAIR.
Resident KV is reported only where the canonical resident probe was run for the composed configuration.
}
\label{tab:reader_quantization}
\begin{tabular}{lccc}
\toprule
Benchmark & PAIR fp16 & PAIR + Reader-int8 & Resident KV \\
\midrule
HotpotQA & 0.7178 & 0.7173 & -- \\
MuSiQue  & 0.4380 & 0.4411 & -- \\
LoCoMo   & 0.4992 & 0.4900 & -- \\
LooGLE   & 0.4438 & 0.4364 & 14.3\% \\
\bottomrule
\end{tabular}
\end{table}

We do not use Reader-int4 as the composition point. On HotpotQA, applying int4 KV quantization to the 7B Reader causes degenerate generation and a severe accuracy collapse, despite int4 remaining effective for the full-context 32B. The failure therefore appears specific to aggressively quantizing the smaller Reader cache rather than to int4 KV quantization in general.

\subsection{Matched-Byte Comparison with Quantized Token Reduction}
\label{app:quantized_matched_byte}

Finally, we ask whether the comparison changes when token-reduction baselines are also combined with quantization. We apply int8 KV quantization to SnapKV and ExpectedAttention and compare them with PAIR using an int8 Reader at similar cache-byte budgets.

For this experiment, we report the batched cache allocation from the accuracy runs divided by the number of original context tokens. This quantity, measured in KiB per original context token, includes padding to the longest row in each batch and should not be interpreted as the batch-one KV memory percentage used in the main accuracy--memory figure.

On HotpotQA, PAIR with an int8 Reader achieves 0.7173 F1 at 46.79 KiB per original context token. SnapKV at a slightly larger budget of 49.92 KiB reaches 0.7023, while ExpectedAttention reaches 0.6436. The lower-budget configurations show the same ordering.

\begin{table}[t]
\centering
\small
\caption{
Matched-byte comparison on HotpotQA.
Cache cost is measured from batched accuracy runs and reported as KiB per original context token.
}
\label{tab:quantized_baselines_hotpot}
\begin{tabular}{lcc}
\toprule
Configuration & KiB / original context token & F1 \\
\midrule
SnapKV 15\% + int8               & 30.30 & 0.6839 \\
ExpectedAttention 15\% + int8   & 31.11 & 0.5521 \\
SnapKV 21.9\% + int8            & 39.62 & 0.6870 \\
ExpectedAttention 21.9\% + int8 & 40.45 & 0.6026 \\
PAIR + Reader-int8              & 46.79 & 0.7173 \\
SnapKV 30\% + int8              & 49.92 & 0.7023 \\
ExpectedAttention 30\% + int8   & 50.72 & 0.6436 \\
Full-context large model + int8                  & 142.45 & 0.7625 \\
Full-context large model fp16                    & 264.64 & 0.7492 \\
\bottomrule
\end{tabular}
\end{table}

The separation is larger on accumulated LooGLE. PAIR with an int8 Reader achieves 0.4364 F1 at 6.90 KiB per original context token, while SnapKV and ExpectedAttention remain substantially less accurate at both smaller and larger nearby budgets.

\begin{table}[t]
\centering
\small
\caption{
Matched-byte comparison on LooGLE.
Cache cost is measured from batched accuracy runs and reported as KiB per original context token.
}
\label{tab:quantized_baselines_loogle}
\begin{tabular}{lcc}
\toprule
Configuration & KiB / original context token & F1 \\
\midrule
SnapKV 15\% + int8               & 4.56  & 0.3261 \\
ExpectedAttention 15\% + int8   & 4.60  & 0.2992 \\
SnapKV 21.9\% + int8            & 6.46  & 0.3589 \\
ExpectedAttention 21.9\% + int8 & 6.47  & 0.3063 \\
PAIR + Reader-int8              & 6.90  & 0.4364 \\
SnapKV 30\% + int8              & 8.65  & 0.3557 \\
ExpectedAttention 30\% + int8   & 8.69  & 0.3276 \\
Full-context large model + int4                  & 14.71 & 0.4828 \\
Full-context large model + int8                  & 27.65 & 0.4927 \\
Full-context large model fp16                    & 53.37 & 0.4876 \\
\bottomrule
\end{tabular}
\end{table}

These matched-byte results support the same conclusion as the resident-memory comparison. Quantization improves the memory efficiency of both PAIR and token-reduction methods, but at comparable cache-byte budgets PAIR retains higher accuracy than the quantized SnapKV and ExpectedAttention configurations evaluated here. The difference is particularly pronounced on accumulated LooGLE, where preserving the complete Reader-side context remains beneficial even after all methods are additionally quantized.

\end{document}